\documentclass[times, preprint, 10pt]{elsarticle}

\usepackage{setspace}
\usepackage{amsmath}                
\usepackage{amssymb}               
\usepackage{amsthm}          
\usepackage{graphicx}             
\usepackage{subcaption}         
\usepackage{circledsteps}           
\usepackage{needspace}         
\usepackage{booktabs}                      
\usepackage{rotating}          
\usepackage{array}
\usepackage{makecell}
\usepackage{multirow}
\usepackage{amssymb}
\usepackage{pdflscape}
\usepackage{hyperref}             
\usepackage[ruled,linesnumbered]{algorithm2e}

\begin{document}

\begin{frontmatter}

\title{UASPL: Uncertainty-Aware Self-Paced Learning with Evidential Neural Networks}

\author[address1]{Yifan Zhang}
\ead{zhangyifannwafu@163.com}

\author[address1]{Yuxin Hu}
\ead{huyuxin@nwafu.edu.cn}

\author[address1]{Zhuobin Hao}
\ead{haozhuobin@126.com}

\author[address1,address2]{Xiaozhuan Gao\corref{cor1}}
\ead{gaoxiaozhuan@nwafu.edu.cn}

\author[address1,address2,address3]{Lipeng Pan\corref{cor1}}
\ead{lipeng.pan@nwafu.edu.cn}

\cortext[cor1]{Corresponding authors}

\affiliation[address1]{College of Information Engineering, Northwest A\&F University, Yangling, 712100, Shaanxi, China}

\affiliation[address2]{Shaanxi Engineering Research Center for Intelligent Perception and Analysis of Agricultural Information, Northwest A\&F University, Shaanxi, China}

\affiliation[address3]{Northwest A\&F University ShenZhen Research Institute, Shenzhen, 518000, China}

\begin{abstract}
Self-paced learning (SPL) is an effective learning paradigm that simulates the human learning process by progressing from easy to difficult samples based on the value of the loss function during the learning process. It has shown great potential in improving model performance and training efficiency. However, the prediction results of samples with smaller loss values are not necessarily reliable, indicating that such samples are not always simple samples for the model. Hence, this article proposes an uncertainty-aware self-paced learning based on evidential neural networks, termed UASPL, which integrates predictive reliability into sample selection through a general loss function within the Subjective Logic framework. This loss function incorporates uncertainty estimation and can be extended to different variants of SPL. Moreover, this loss function couples a sample selection preference, thereby ensuring the interpretability of the sample selection process. Finally, the experimental results on multiple datasets show that UASPL outperforms other SPL methods in terms of classification performance, interpretability, and generality. The source code is available at: \href{https://github.com/treelife979/UASPL}{https://github.com/treelife979/UASPL}.
\end{abstract}


\end{frontmatter}

\section{Introduction}
\label{introduction}

Self-paced learning (SPL) \cite{kumar2010self}, rooted in Curriculum Learning (CL) \cite{bengio2009curriculum}, dynamically selects training data in an easy-to-hard manner rather than indiscriminately using all training data as in traditional machine learning and deep learning methods, thereby achieving remarkable success in robustness and training efficiency \cite{li2022unsupervised, yang2024advancing, zhao2024symmetric}.
Extensions of SPL mainly improve the framework from two aspects: one focusing on the mathematical design of the self-paced regularization term to adjust sample weights \cite{jiang2014easy, zhao2015matrix, li2018ML}, and the other involving the introduction of prior knowledge to characterize sample difficulty \cite{jiang2015SPCL, GUO_2018, zhang2024weighted}. For the former, anchored in theoretical deduction and formal generalization, it breaks the limitations of traditional regularization terms (e.g., hard/linear forms) by extending their mathematical structures. For the latter, it incorporates prior knowledge into the weight constraints, correcting the distributional biases that are solely based on the loss. These developments have substantially enriched the SPL framework.

\begin{figure}[!ht]
  \centering
  
\begin{subfigure}[b]{0.48\textwidth}
  \includegraphics[width=\linewidth]{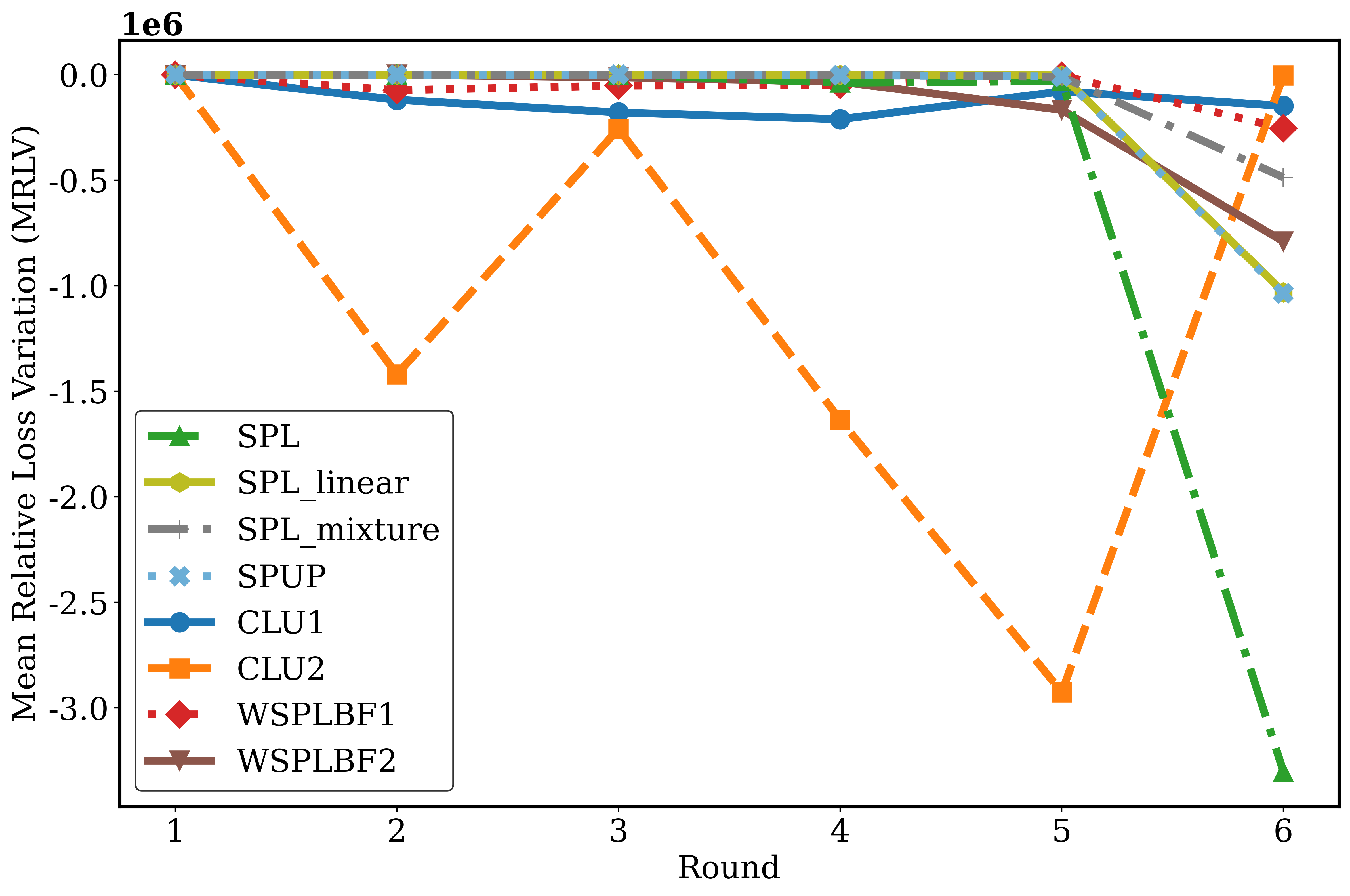}
  \caption{Transfusion}
\end{subfigure}
\hfill
\begin{subfigure}[b]{0.48\textwidth}
  \includegraphics[width=\linewidth]{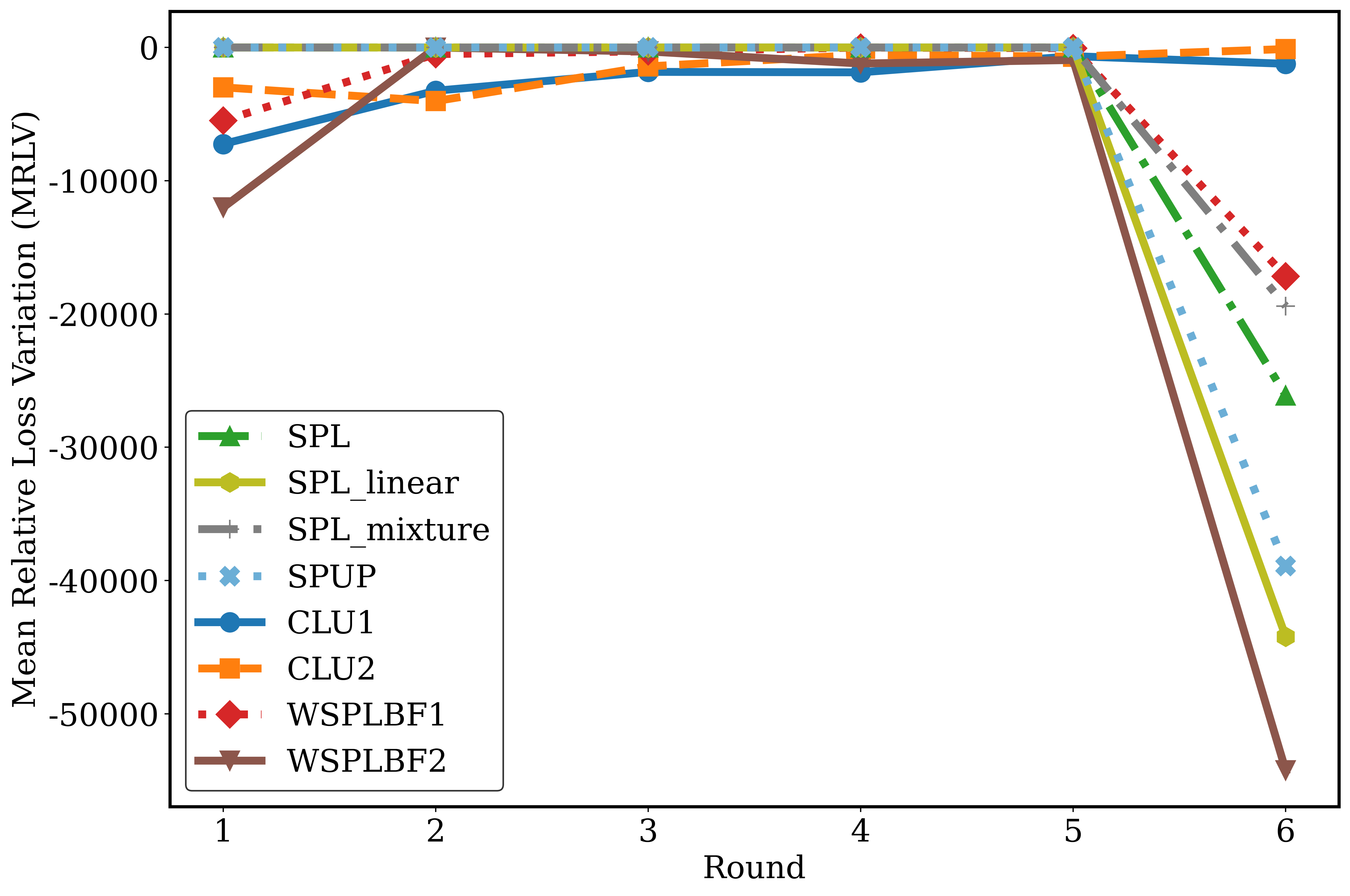}
  \caption{Pop}
\end{subfigure}
\vspace{0.5em}
\begin{subfigure}[b]{0.48\textwidth}
  \includegraphics[width=\linewidth]{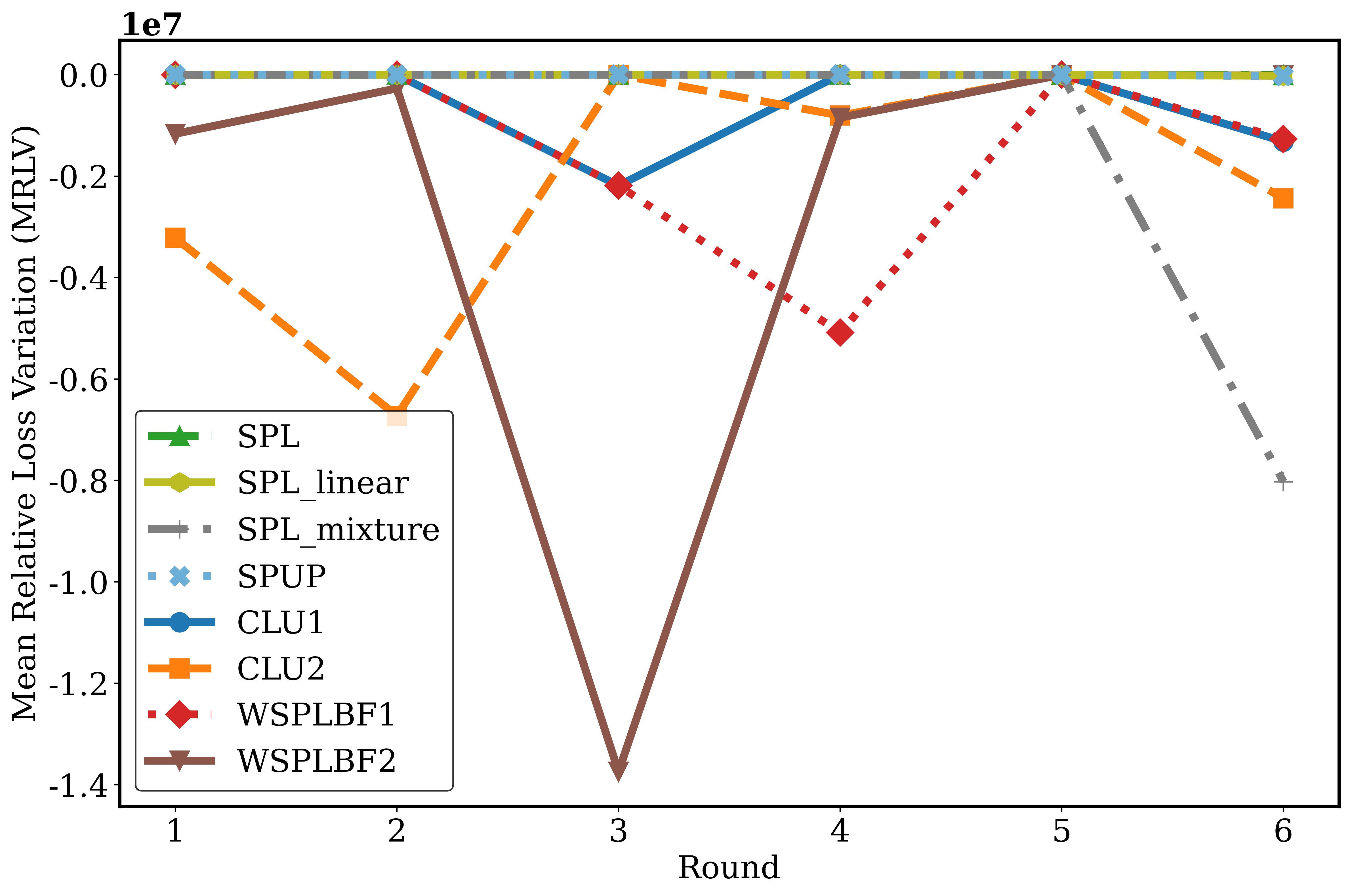}
  \caption{Breast\_cancer\_Diagnostic}
\end{subfigure}
\hfill
\begin{subfigure}[b]{0.48\textwidth}
  \includegraphics[width=\linewidth]{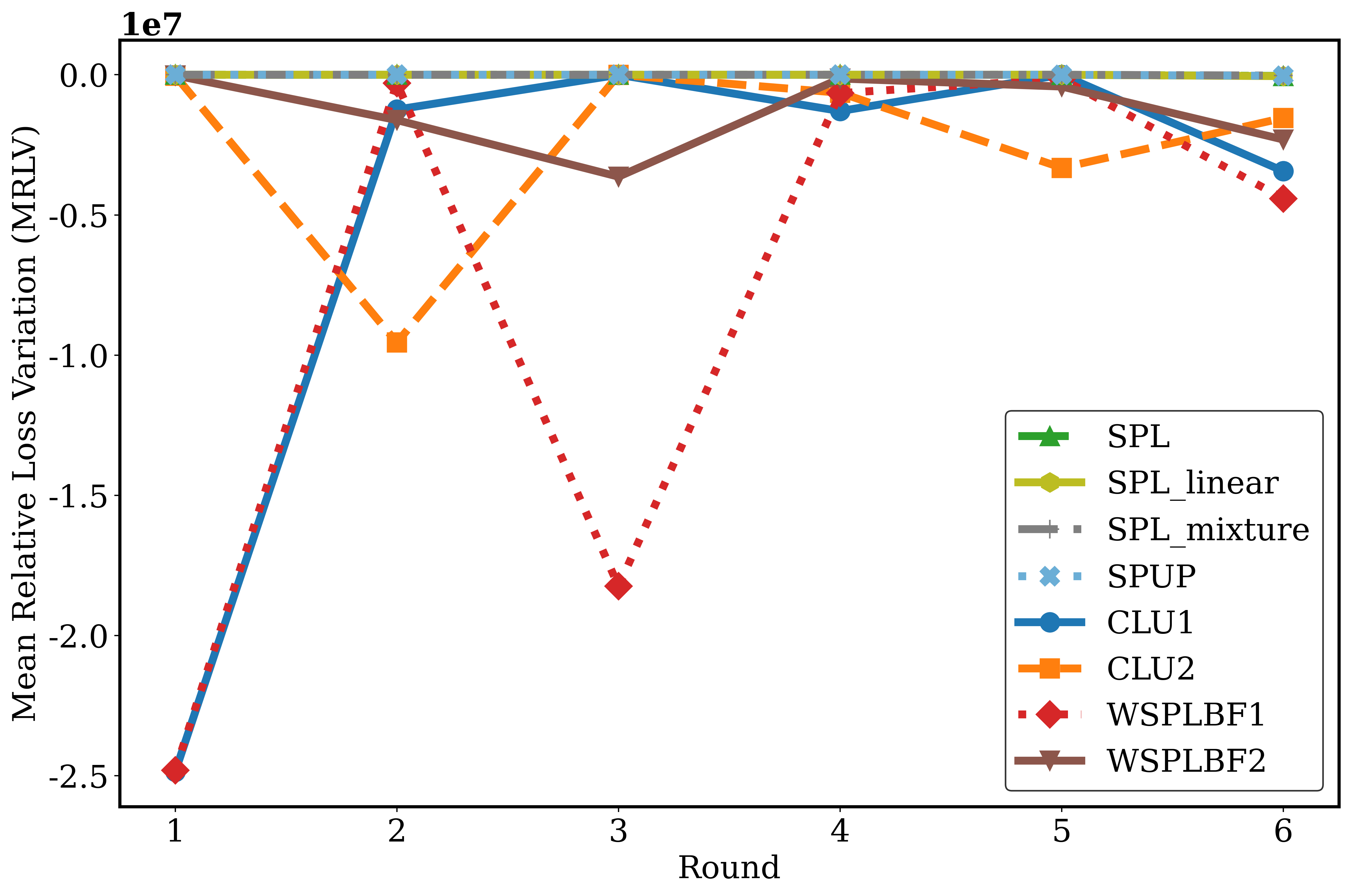}
  \caption{Ionosphere}
\end{subfigure}

\caption{Mean relative loss variation across rounds on four datasets.}
\label{fig:gain_ratio}

\par
\vspace{5pt}
\begin{minipage}{\textwidth}
\raggedright
\footnotesize
\textit{Note:} 
Let $S_1$ denote the set of samples selected in the first round. For each round $t$, the relative loss variation of a sample $x \in S_1$ is defined as $\frac{L_{t-1}(x)-L_t(x)}{L_{t-1}(x)+\epsilon}$, where $L_t(x)$ denotes the loss of sample $x$ after round $t$, $L_0(x)$ denotes the loss obtained by a forward pass before the first round of self-paced learning begins, and $\epsilon$ is a small positive constant used to avoid division by zero. Each curve reports the mean value of this quantity over all samples in $S_1$.

\end{minipage}
\end{figure}

However, a pivotal issue remains insufficiently explored: Are low-loss samples reliably easy samples that the model requires? Intuitively, the mean relative loss variation (MRLV) of simple samples selected in the first round should remain relatively stable in subsequent rounds and should not experience a significant decline. The experimental result of Fig.~\ref{fig:gain_ratio} contradict this intuition on multiple datasets. As shown in the Fig.~\ref{fig:gain_ratio}, the MRLV fluctuates or decreases significantly in the subsequent training stages, and even becomes negative, which reflects that their losses are increasing in the training stages thereafter. This anomaly indicates that the samples selected based on the losses in the first round are not reliably easy samples for all stages of the model.

To better distinguish reliably simple samples from pseudo-easy ones, this article proposes UASPL, an uncertainty-aware self-paced learning approach based on evidential neural networks within the framework of Subjective Logic. Specifically, UASPL incorporates evidential uncertainty generated by the current model together with the label-fitting loss, so that sample-difficulty estimation no longer depends on training loss alone and does not require prior knowledge. By adaptively guiding the growth of evidence, UASPL further induces an interpretable sample-selection preference consistent with the easy-to-hard learning strategy. Moreover, this design can also be naturally extended to different SPL variants without substantial modification. Experimental results on multiple datasets show that UASPL achieves favorable classification performance, interpretability, and generality compared with representative SPL methods. In conclusion, the contributions of this article are summarized as follows:

(1) To the best of our knowledge, UASPL is the first self-paced learning method to directly incorporate model-generated evidential uncertainty together with the label-fitting loss into the self-paced learning objective, ensuring that it can select reliably simple samples.

(2) The insights into why UASPL works can be well interpreted. The sample-selection preference induced by UASPL is consistent with the self-paced learning principle that reliable easy samples should be selected earlier, while samples with insufficient or misleading evidence should be delayed.

(3) The generality of UASPL beyond a specific SPL formulation is analyzed. Particularly, UASPL is embedded into different SPL variants, such as the linear and mixture regularizers, to assess whether UASPL is merely tied to a particular regularizer.

The rest of this paper is organized as follows. Section~\ref{sec_related} reviews the related work. Section~\ref{sec_uaspl} elaborates on UASPL, including the formulation of the loss function, the corresponding principle and the detailed algorithm. Section ~\ref{sec_experiments} presents and analyzes the experimental results on multiple datasets, demonstrating the effectiveness of the proposed method. Finally, Section~\ref{sec_conclusion} concludes this paper and outlines the future work.

\section{Related Work}
\label{sec_related}

This section reviews prior studies most relevant to our method, including uncertainty estimation in deep learning and self-paced learning.

\subsection{Uncertainty Estimation in Deep Learning}

Uncertainty estimation has been widely studied in deep learning because it provides reliability information regarding the prediction results of the model. One common line of work derives uncertainty-related signals directly from model outputs, such as confidence \cite{chang2017active}, margin \cite{islam2023paced}, and the variance of historical predictive probabilities \cite{feng2025your}. Another influential direction is Bayesian methods. In this line, Monte Carlo dropout performs multiple stochastic forward passes to approximate Bayesian inference \cite{gal2016dropout}, from which uncertainty measures such as mutual information \cite{gal2017deep} and its variants \cite{kim2021task,woo2023active} can be estimated.

Different from these approaches, evidential deep learning (EDL) quantifies predictive uncertainty through an evidential neural network (ENN) \cite{sensoy2018evidential, zhang2025information}, which parameterizes a higher-order Dirichlet distribution over class probabilities within the framework of Subjective Logic \cite{dong2024novel, zhang2025survey}, thereby alleviating the overconfidence issue of traditional deep neural networks \cite{chen2025revisiting, wang2025regularized}. Under this formulation, the ENN outputs evidence for each class, from which both predictive probabilities and uncertainty estimates can be derived \cite{wang2022uncertainty, gao2025comprehensive}. One commonly used loss function in EDL is the Evidential Mean Square Error (EMSE), formulated as:
\begin{equation}\label{emse}
\mathcal{L}_{\text{EMSE}} = \sum_{k=1}^{K} \left( y_k - \frac{\alpha_k}{S} \right)^2 + \frac{\alpha_k (S - \alpha_k)}{S^2 (S + 1)}
\end{equation}
where $y_k$ is the one-hot encoded ground truth and $\alpha_k = e_k + 1$. $e_k$ is the evidence of the \textit{k}-th class, and $S = \sum_{j=1}^{K} \alpha_j$. To further guide the learning process, a Kullback–Leibler (KL) divergence term is introduced to penalize evidence accumulation on non-ground-truth classes, defined as:
\begin{equation}\label{kl}
    \mathcal{L}_{\text{KL}} = \text{KL} \left[ \text{Dir}(\tilde{\alpha}) \parallel \text{Dir}(1, \ldots, 1) \right]
\end{equation}
where $\tilde{\alpha}_i = \mathbf{y}_i + (1 - \mathbf{y}_i) \odot \alpha_i$, and then the final loss function is described as follows:
\begin{equation}\label{tag:KL}
    \mathcal{L}_{EDL} = \mathcal{L}_{\text{EMSE}} + \lambda_{KL} \cdot \mathcal{L}_{\text{KL}}
\end{equation}
where $\lambda_{KL}$ is an annealing coefficient. Furthermore, uncertainty estimation $u$ is used to assess the reliability of predictions, defined as:
\begin{equation}\label{tag:u}
u=\frac{K}{S}
\end{equation}

\subsection{Self-Paced Learning}
\label{subsec2.1}

Self-paced learning is a machine learning paradigm that simulates the human learning process by gradually introducing training samples from easy to difficult \cite{kumar2010self, meng2017theoretical, kang2023self}. Specifically, SPL assigns a weight to each sample, prioritizing easy samples at early stages and progressively incorporating more difficult ones \cite{yan2023bi, kang2024fine, poyser2026dds}. The optimization objective of SPL can be expressed as
\begin{equation}
\min _{w, v} \sum_{i=1}^{N} v_{i} L\left(y_{i}, f\left(x_{i} ; w\right)\right)+ g\left(v_{i} ; \lambda\right)
\end{equation}
where $x_i$ and $y_i$ are the feature and label of the \textit{i}-th sample, respectively, $f(x_i;w)$ is the learned model with model parameter $w$, $L(\cdot)$ is the loss function, $v_i$ is the sample weight controlling the learning order, and $g(v_i;\lambda)$ is the self-paced regularizer. More details of SPL can be found in \cite{kumar2010self,jiang2014easy,zhao2015matrix}.

Studies on SPL have mainly been conducted from the following two perspectives: how to update sample weights and how to characterize sample difficulty during progressive learning.
One important line of work redesigns self-paced regularizers to obtain more flexible weighting behavior in different learning scenarios. Representative studies have introduced multiple soft regularizers for self-paced reranking \cite{jiang2014easy}, extended SPL to matrix factorization \cite{zhao2015matrix}, developed general self-paced functions for multi-label learning \cite{li2018ML}, and proposed dynamic or adaptive weighting strategies for convolutional or cost-sensitive settings \cite{li2017spcn, lihao2021cost}. Although these studies improve the flexibility of the SPL framework, sample selection still relies primarily on loss-based criteria, making it difficult to select truly simple samples, as shown in Fig.~\ref{fig:gain_ratio}.

Another popular line of work incorporates additional information beyond the training loss for sample-difficulty estimation. For example, Self-Paced Learning with Statistics Uncertainty Prior (SPUP) introduces a statistics uncertainty prior into SPL, which  alleviates the bias of purely loss-based selection \cite{GUO_2018}. Weighted self-paced learning with belief functions (WSPLBF) combines learning loss with belief-function-based evidential uncertainty to refine sample selection \cite{zhang2024weighted}. 
Although these studies suggest that using uncertainty as an additional prior can improve sample-difficulty estimation, they differ essentially from UASPL, as shown in Table~\ref{tab:related_spl_comparison}.

\begin{table}[!ht]
\centering
\caption{Comparison between UASPL and prior self-paced learning methods.}
\label{tab:related_spl_comparison}
\renewcommand{\arraystretch}{1.1}
\begin{tabular}{c c c c c}
\hline
\multirow{2}{*}{Method} & 
\multirow{2}{*}{\begin{tabular}[c]{@{}c@{}}Dirichlet\\parameters\end{tabular}} & 
\multicolumn{3}{c}{Uncertainty} \\
\cline{3-5}
 & & Internal & Dynamic & \begin{tabular}[c]{@{}c@{}}Selection\\criterion\end{tabular} \\
\hline
Traditional SPL & $\times$ & $\times$ & $\times$ & $\times$ \\
SPUP & $\times$ & $\times$ & \checkmark & $\times$ \\
WSPLBF & $\times$ & $\times$ & $\times$ & \checkmark \\
UASPL & \checkmark & \checkmark & \checkmark & \checkmark \\
\hline
\end{tabular}
\end{table}

Table~\ref{tab:related_spl_comparison} summarizes the main distinctions between UASPL and prior self-paced learning methods from the following four perspectives:
(1) Compared with traditional SPL, SPUP, and WSPLBF, UASPL parameterizes a Dirichlet distribution over class probabilities rather than point-estimate probabilities, enabling predictive uncertainty quantification. 
(2) Uncertainty in UASPL is generated internally by the current evidential model, whereas SPUP relies on a predefined Gaussian perturbation over latent sample weights and WSPLBF constructs uncertainty externally from belief functions. 
(3) Uncertainty in UASPL evolves with model-parameter updates during training. In contrast, the uncertainty in WSPLBF remains fixed with respect to the learned classifier, while the dynamics in SPUP mainly comes from the pace-dependent uncertainty prior rather than model-generated uncertainty. 
(4) Although WSPLBF also uses uncertainty in its sample-selection criterion, UASPL differs in that its selection preference is shaped by uncertainty estimated from the current model state. 
Furthermore, as illustrated in Fig.~\ref{fig:gain_ratio}, both SPUP and WSPLBF still have room for improvement in selecting truly easy samples.

\section{Uncertainty-Aware Self-Paced Learning}
\label{sec_uaspl}

In order to select truly simple samples based on the loss value, this section details the UASPL, including its loss function, theoretical principle and detailed algorithmic implementation. 

\subsection{UASPL Model}
\label{subsec3.1}

Uncertainty estimation (or reliability estimation) quantifies the prediction confidence of the model for the sample, reflecting the learning ability of current model \cite{xu2025Interactive,liu2024multisource}. Nevertheless, existing self-paced learning methods only focus on the label fitting loss of sample to measure sample difficulty during the training process, while neglecting uncertainty estimation of predictions, making it challenging to select truly simple samples. Thus, this subsection designs a general loss function, which incorporates the reliability of predictions into the sample selection strategy based on the loss values of samples, facilitating the selection of reliably simple samples. The specific form is elaborated below.

Suppose that $x = [x_1,...,x_n]$ is the training samples, $y = [y_1, ..., y_n]$ is the labels for corresponding samples. Accordingly $v = [v_1,...,v_n]^T$ denotes a vector of weight variables for each training sample and \textit{w} is the model parameter. Given an input sample $x_i$, the EDL model outputs nonnegative evidence values $e_i=[e_{i1},e_{i2},\ldots,e_{iK}]$ and the corresponding Dirichlet parameters $\alpha_i=[\alpha_{i1},\alpha_{i2},\ldots,\alpha_{iK}]$ for $K$ classes. Following the EDL notation introduced in Section~\ref{sec_related}, we have $\alpha_{ij}=e_{ij}+1$, $S_i=\sum_{j=1}^{K}\alpha_{ij}$, and $u_i=K/S_i$. Therefore, the learning objective of UASPL, which integrates both uncertainty estimation and loss values of samples, is formulated as follows:
\begin{align}
\label{tag:L}
\mathcal{L}(x,y;w,v)
&=
\sum_{i=1}^{n} v_i \mathcal{L}_{\text{total}}^{(i)}(x,y;w)
-
\lambda \sum_{i=1}^{n} v_i \nonumber \\
&=
\sum_{i=1}^{n} v_i
\left(
\mathcal{L}_{\text{EMSE}}^{(i)}(x,y;w)
+
\left((1-c_i)(1-u_i)+c_i u_i\right)
\mathcal{L}_{\text{KL}}^{(i)}(x,y;w)
\right)
-
\lambda \sum_{i=1}^{n} v_i ,
\end{align}
where $\lambda$ is the age parameter in the self-paced hard regularizer. In Eq.~\ref{tag:L}, $\mathcal{L}_{\text{total}}^{(i)}$ denotes the uncertainty-aware loss of the $i$-th sample before being weighted by variable $v_i$. It combines the label-fitting term and the uncertainty-weighted KL term, and is formulated as
\begin{equation}
\label{tag:L_total_def}
\mathcal{L}_{\text{total}}^{(i)}
=
\mathcal{L}_{\text{EMSE}}^{(i)}
+
\text{coeff}_i \mathcal{L}_{\text{KL}}^{(i)},
\end{equation}
where $\mathcal{L}_{\text{EMSE}}^{(i)}$ is the Evidential Mean Square Error, $\mathcal{L}_{\text{KL}}^{(i)}$ is the Kullback–Leibler divergence term, and the coefficient $\text{coeff}_i$ is given by
\begin{equation}
\text{coeff}_i =
\begin{cases}
u_i, & \text{if } c_i = 1, \\[4pt]
1-u_i, & \text{if } c_i = 0,
\end{cases}
\end{equation}
which can be compactly rewritten as $\text{coeff}_i = (1-c_i)(1-u_i) + c_i u_i$. Here, $u_i$ denotes the evidential uncertainty of the $i$-th sample, and $c_i$ is the prediction correctness indicator, with $c_i=1$ for a correct prediction and $c_i=0$ otherwise.
The coefficient is designed to adaptively adjust the inhibitory effect of the KL term on non-target-label evidence according to both evidence sufficiency and prediction correctness. For correctly predicted samples, higher uncertainty indicates insufficient supporting evidence, so setting \(\text{coeff}_i=u_i\) strengthens the KL effect and guides reliable evidence growth by inhibiting non-target-label evidence. For incorrectly predicted samples, lower uncertainty indicates confident but misleading evidence, so setting \(\text{coeff}_i=1-u_i\) strengthens the KL effect and suppresses erroneous evidence accumulation.

\subsection{UASPL Principle and Algorithm}
\label{subsec3.2}

The previous subsection presents a loss function that aggregates the uncertainty estimation and the label-fitting loss. This subsection will analyze a sample selection preference guided by the loss function. 
For the subsequent analysis, let $m$ denote an arbitrary class label of the $i$-th sample, while $l$ and $k$ denote the target label and the non-target label of the $i$-th sample, respectively.

In Eq.~\ref{tag:L}, $\mathcal{L}_{\text{total}}^{(i)}$ is weighted by $v_i$, while the self-paced regularization term $-\lambda\sum_{r=1}^{n}v_r$ is independent of the Dirichlet parameters. Hence, taking the derivative of $\mathcal{L}$ with respect to $\alpha_{im}$ gives
\begin{equation} 
\frac{\partial \mathcal{L}}{\partial \alpha_{im}}
=
v_i
\frac{\partial \mathcal{L}_{\text{total}}^{(i)}}{\partial \alpha_{im}}
+
\frac{\partial}{\partial \alpha_{im}}
\left(
-\lambda \sum_{r=1}^{n} v_r
\right)
=
v_i
\frac{\partial \mathcal{L}_{\text{total}}^{(i)}}{\partial \alpha_{im}}.
\end{equation}
Obviously, the derivative of $\mathcal{L}$ with respect to $\alpha_{im}$ is given by the weighted sample-wise derivative $v_i \frac{\partial \mathcal{L}_{\text{total}}^{(i)}}{\partial \alpha_{im}}$. Here, $v_i$ only scales the sample-wise gradient and does not change the preference determined by $\mathcal{L}_{\text{total}}^{(i)}$. Thus, to analyze the sample-selection preference induced by UASPL, we focus on
$\frac{\partial \mathcal{L}_{\text{total}}^{(i)}}{\partial \alpha_{im}}$. According to Eq.~\ref{tag:L_total_def}, the derivative of $\mathcal{L}_{\text{total}}^{(i)}$ with respect to $\alpha_{im}$ is

\begin{equation}
\label{tag:partial_L}
\frac{\partial \mathcal{L}_{\text{total}}^{(i)}}{\partial \alpha_{im}}
=
\frac{\partial \mathcal{L}_{\text{EMSE}}^{(i)}}{\partial \alpha_{im}}
+
\text{coeff}_i
\frac{\partial \mathcal{L}_{\text{KL}}^{(i)}}{\partial \alpha_{im}}
+
\mathcal{L}_{\text{KL}}^{(i)}
\frac{\partial \text{coeff}_i}{\partial \alpha_{im}} .
\end{equation}

\textbf{(1) Analysis of the first term in Eq.~\ref{tag:partial_L}}. The formula for the EMSE is presented as follows:
\begin{align}
\label{tag:derivation}
\mathcal{L}_{EMSE}^{(i)}
&=
\sum_{j=1}^{K}
\left[
\left(y_{ij}-E[p_{ij}]\right)^2
+
\mathrm{Var}(p_{ij})
\right] \nonumber\\
&=
\sum_{j=1}^{K}
\underbrace{\left(y_{ij}-\frac{\alpha_{ij}}{S_i}\right)^2}_{\mathcal{L}_{ij}^{err}}
+
\underbrace{
\frac{\alpha_{ij}(S_i-\alpha_{ij})}{S_i^2(S_i+1)}
}_{\mathcal{L}_{ij}^{var}} \nonumber\\
&=
\sum_{j=1}^{K}
\left[
\left(y_{ij}-\mu_{ij}\right)^2
+
\sigma_{ij}^{2}
\right],
\end{align}
To facilitate the subsequent analysis, the results of taking the partial derivatives of $\mu_{ij}$ and $\sigma_{ij}$ in Eq.~\ref{tag:derivation} with respect to the Dirichlet parameter $\alpha_{im}$ (where \textit{m} represents any label of the $i$-th sample) are displayed as follows:
\begin{equation}
\frac{\partial \mu _{ij}}{\partial \alpha  _{im}}=\frac{S_{i}-\alpha _{ij}}{S_{i}^{2}}\left(> 0,j=m \right) \quad \frac{\partial \mu _{ij}}{\partial \alpha  _{im}}=-\frac{\alpha _{ij}}{S_{i}^{2}}\left ( < 0,j\neq m \right )
\end{equation}
\begin{equation}
 \frac{\partial \sigma_{ij}^{2}}{\partial \alpha_{im}}=\frac{\left ( 1-2\mu _{ij} \right )}{S_{i}+1}\frac{\partial \mu _{ij}}{\partial \alpha _{im}}-\frac{ \mu _{ij}\left (1-\mu _{ij}  \right)}{\left (S_{i} + 1 \right )^{2}} \underbrace{\frac{\partial S_{i}}{\partial \alpha _{im}}}_{=1}
 \end{equation}

 Next, we sequentially analyze the partial derivatives of \(\mathcal{L}_{EMSE}^{(i)}\) and \(\mathcal{L}_{KL}^{(i)}\) with respect to \(\alpha_{il}\) and \(\alpha_{ik}\).

\Circled{1} For the target label $l$ of the $i$-th sample, the partial derivative of $\mathcal{L}_{EMSE}$ with respect to the Dirichlet parameter $\alpha_{il}$ can be calculated as follows:
\begin{align}\label{tag:emse_il}
 \frac{\partial \mathcal{L}_{EMSE}^{(i)}}{\partial \alpha_{il}}
 &=\sum_{j=1}^{K} \frac{\partial \mathcal{L} _{ij}^{err}}{\partial \alpha_{il}} + \frac{\partial \mathcal{L}_{ij}^{var}}{\partial \alpha_{il}} \nonumber\\
 &=\sum_{j=1}^{K} \left(-2 \right )\left ( y_{ij}-\mu _{ij} \right ) \frac{\partial \mu_{ij}}{\partial \alpha_{il}}
 +\left [ \left( \frac{1-2\mu_{ij}}{S_{i}+1}\right)\frac{\partial \mu _{ij}}{\partial \alpha_{il}} -\frac{\mu_{ij}\left( 1-\mu_{ij} \right)}{\left( S_{i}+1\right)^{2}}\right]
\end{align}

When $j=l$ and $j \neq l$ in the prediction error term $\left(-2 \right )\left( y_{ij}-\mu _{ij} \right ) \frac{\partial \mu_{ij}}{\partial \alpha_{il}}$ in Eq.~\ref{tag:emse_il}, the partial derivative of the prediction error term is respectively exhibited as follows:
\begin{equation}\label{tag:both}
    -2\left(1-\mu_{il}\right) \underbrace{\frac{\partial \mu_{il}}{\partial \alpha_{il}}}_{>0} < 0, \quad -2\sum_{j=1,j\neq l}^{K} \left(- \mu_{ij} \right) \underbrace{\frac{\partial \mu_{ij}}{\partial \alpha_{il}}}_{< 0} < 0
\end{equation}
It can be seen from Eq.~\ref{tag:both} that the partial derivative of the prediction error term in Eq.~\ref{tag:emse_il} is negative for both $j=l$ and $j \neq l$. In the derivative of the predicted variance term, for the second term, $ -\frac{\mu_{ij}\left( 1-\mu_{ij} \right)}{\left( S_{i}+1\right)^{2}}<0$ because of $\mu_{ij}\in \left ( 0,1 \right )$. For the first term in the predicted variance term, its value depends on the value of $\mu_{ij}$ and whether $j$ is equal to $l$, which can be either greater than 0 or less than 0. Hence, for a definite judgment, $\frac{\partial \mathcal{L}_{EMSE}^{(i)}}{\partial \alpha_{il}}$ is described as follows:
\begin{equation}\label{tag:2_detail}
    \frac{\partial \mathcal{L}_{EMSE}^{(i)}}{\partial \alpha_{il}}
     =(\frac{\partial \mathcal{L} _{il}^{err}}{\partial \alpha_{il}} + \frac{\partial \mathcal{L}_{il}^{var}}{\partial \alpha_{il}})
     + \sum_{j=1,j\neq l}^{K} (\frac{\partial \mathcal{L} _{ij}^{err}}{\partial \alpha_{il}} + \frac{\partial \mathcal{L}_{ij}^{var}}{\partial \alpha_{il}})
\end{equation}

For the target label $l$ and the non-target labels:
\begin{equation*}
\frac{\partial \mathcal{L} _{il}^{err}}{\partial \alpha_{il}} + \frac{\partial \mathcal{L}_{il}^{var}}{\partial \alpha_{il}}
    =\underbrace{\frac{2S_{i}\left(\mu_{il}-1\right)-1}{S_{i}+1} \cdot \frac{1-\mu_{il}}{S_{i}}}_{< 0}\underbrace{-\frac{\mu_{il}\left(1-\mu_{il}\right)}{\left(S_{i}+1\right)^2}}_{< 0}< 0
\end{equation*}

\begin{equation*}
\sum_{j=1,j\neq l}^{K}(\frac{\partial \mathcal{L} _{ij}^{err}}{\partial \alpha_{il}} + \frac{\partial \mathcal{L}_{ij}^{var}}{\partial \alpha_{il}})
    =\sum_{j=1,j\neq l}^{K} (\underbrace{\frac{-\mu_{ij}\left(2S_{ij}\mu_{ij}+1\right)}{(S_{i}+1)S_{i}}}_{< 0}\underbrace{-\frac{\mu_{ij}\left(1-\mu_{ij}\right)}{\left(S_{i}+1\right)^2}}_{< 0})< 0
\end{equation*}

Thus, $\frac{\partial \mathcal{L}_{EMSE}^{(i)}}{\partial \alpha_{il}} < 0$.

\Circled{2} For the non-target label $k$ of the $i$-th sample, a general result of the partial derivative of the $\mathcal{L}_{EMSE}^{(i)}$ with respect to the Dirichlet parameter $\alpha_{ik}$ is calculated as follows: $\frac{\partial \mathcal{L}_{EMSE}^{(i)}}{\partial \alpha_{ik}} =  \frac{2\alpha_{il}}{S^{2}}-
    \frac{2\left(S-\alpha_{ik}\right)}{S\left(S+1\right)}+\sum_{j=1}^{K}\frac{\left(2S+1\right)\alpha_{ij}\left(S-\alpha_{ij}\right)}{\left(S^{2}+S\right)^{2}}$.
For the sake of comparison, the expression in Eq.~\ref{tag:emse_il} can be written as: $\frac{\partial \mathcal{L}_{EMSE}^{(i)}}{\partial \alpha_{il}} =  \frac{2\alpha_{il}}{S^{2}}- \frac{2}{S}-
    \frac{2\left(S-\alpha_{il}\right)}{S\left(S+1\right)}+\sum_{j=1}^{K}\frac{\left(2S+1\right)\alpha_{ij}\left(S-\alpha_{ij}\right)}{\left(S^{2}+S\right)^{2}}$. 
Obviously, when $\alpha_{ik}=\alpha_{il}$, $\frac{\partial \mathcal{L}_{EMSE}^{(i)}}{\partial \alpha_{il}} < \frac{\partial \mathcal{L}_{EMSE}^{(i)}}{\partial \alpha_{ik}}$. $\frac{\partial \mathcal{L}_{EMSE}^{(i)}}{\partial \alpha_{il}}$ is definitely less than 0, while $\frac{\partial \mathcal{L}_{EMSE}^{(i)}}{\partial \alpha_{ik}}$ can be either positive or negative. Thus, the model's driving effect on the increase of evidence for non-target labels is definitely smaller than that on the target label.

\textbf{(2) Analysis of the second term in Eq.~\ref{tag:partial_L}}. For the target label $l$ of the $i$-th sample, it is obvious that the KL divergence term has a derivative of 0 with respect to $\alpha_{il}$, i.e., $\frac{\partial \mathcal{L}_{\text{KL}}^{(i)}}{\partial \alpha_{il}}=0$. Moreover, for the non-target label $k$ of the $i$-th sample,  the partial derivative of KL divergence term with respect to $\alpha_{ik}$ is shown as follows:
\begin{equation}\label{tag:kl_ik}
     \frac{\partial \mathcal{L}_{\text{KL}}^{(i)}}{\partial \alpha_{ik}}
     = \left ( \alpha_{ik}-1 \right ){\psi }'\left ( \alpha_{ik} \right )-\left ( \widetilde{S}_{i}-K \right ){\psi }'\left ( \widetilde{S}_{i} \right )
\end{equation}
where $\widetilde{S_{i}}=\sum_{j=1,j \neq l}^{K}\alpha_{ij}+1$. 
Based on the properties and limit behavior of the Trigamma function, when the evidence is insufficient ($ S\rightarrow K, \alpha_{ik} \rightarrow 1$), the value of Eq.~\ref{tag:kl_ik} tends to be greater than 0. When the evidence is sufficient ($ S \rightarrow \infty $), the value of Eq.~\ref{tag:kl_ik} needs to be discussed in the following two cases. For one, the value of Eq.~\ref{tag:kl_ik} tends to 0 when the evidence is evenly distributed. For another, when the evidence is unevenly distributed, the partial derivative corresponding to the non-target label with the maximum evidence is positive (pushing the evidence to contract and suppressing excessive accumulation), while the partial derivatives corresponding to the other non-target labels are negative (pushing the evidence to grow relatively and calibrating the uniformity of the distribution).

In addition, the $\text{coeff}_i$, which incorporates the prediction correctness indicator $c_i$ and uncertainty $u_i$, only scales the value of \(\frac{\partial \mathcal{L}_{KL}^{(i)}}{\partial \alpha_{im}}\) rather than change the sign of \(\frac{\partial \mathcal{L}_{KL}^{(i)}}{\partial \alpha_{im}}\). Specifically, when samples are correctly predicted ($\text{coeff}_i = u_i$), increased uncertainty enhances the regulatory effect of the KL divergence term. In contrast, when samples are incorrectly predicted ($\text{coeff}_i =1- u_i$), reduced uncertainty strengthens this regulatory effect.

\textbf{(3) Analysis of the third term in Eq.~\ref{tag:partial_L}}. The third term in Eq.~\ref{tag:partial_L} is computed as

\begin{equation}
\mathcal{L}_{\text{KL}}^{(i)}
\frac{\partial \text{coeff}_i}{\partial \alpha_{im}} \nonumber
=
\mathcal{L}_{\text{KL}}^{(i)}
(2c_i-1)\left(-\frac{K}{S_i^2}\right).
\end{equation}
For the target label $l$ or the non-target label $k$ of the $i$-th sample, the value of the third term in Eq.~\ref{tag:partial_L} is the same, which does not affect the relative magnitudes of $\partial \mathcal{L}_{\text{total}}^{(i)}/\partial \alpha_{il}$ and $\partial \mathcal{L}_{\text{total}}^{(i)}/\partial \alpha_{ik}$. Furthermore, it adjusts the total evidence strength according to prediction correctness. That is to say, when $c_i=1$, the third term is negative, which increases the Dirichlet parameters under gradient descent and encourages evidence accumulation. When $c_i=0$, the third term is positive, which decreases the Dirichlet parameters under gradient descent and suppresses evidence accumulation.

\textbf{(4) Interpretable Sample Selection Preference in UASPL.}

Based on the above analysis, the sample-selection preference induced by UASPL can be revealed by sorting $\mathcal{L}_{\text{total}}^{(i)}$ in ascending order, yielding the following easier-to-harder sequence:

\needspace{4\baselineskip}
\begin{itemize}
    \item correct prediction with low uncertainty
    \item correct prediction with high uncertainty
    \item incorrect prediction with high uncertainty
    \item incorrect prediction with low uncertainty
\end{itemize}

This ordering can be understood from the roles of prediction correctness and evidential uncertainty in $\mathcal{L}_{\text{total}}^{(i)}$. Correctly predicted samples usually have smaller EMSE loss than incorrectly predicted ones. For correctly predicted samples, where $\text{coeff}_i=u_i$, lower uncertainty reduces the contribution of the KL term and leads to a smaller $\mathcal{L}_{\text{total}}^{(i)}$. For incorrectly predicted samples, where $\text{coeff}_i=1-u_i$, higher uncertainty reduces the contribution of the KL term, whereas low uncertainty indicates misleading evidence and results in a larger $\mathcal{L}_{\text{total}}^{(i)}$.

\begin{figure}[!ht]
  \centering
  \includegraphics[width=0.95\textwidth]{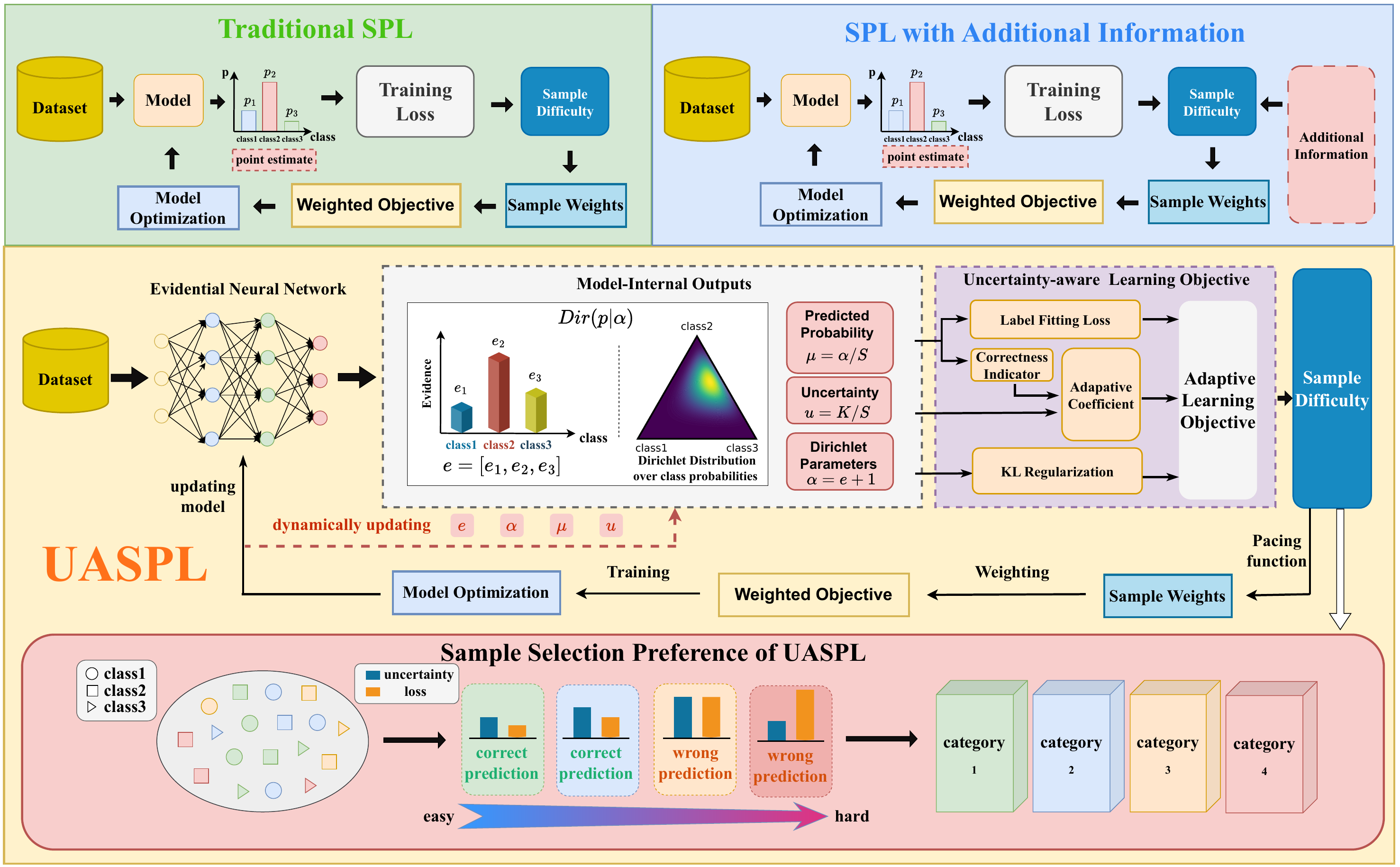}
  \caption{Framework of UASPL and comparison with prior SPL methods. Traditional SPL estimates sample difficulty mainly according to training loss. SPL with additional information further incorporates external auxiliary information into the difficulty measure. 
  In contrast, UASPL integrates model-generated evidential uncertainty and the label-fitting loss into the self-paced learning objective to characterize sample difficulty, thereby enabling reliability-aware selection.
  The bottom part illustrates the sample selection preference of UASPL from easy to hard.}
  \label{img:framework}
\end{figure}

For these four categories of samples with progressively increasing difficulty levels, UASPL produces different evidence-adjustment effects, as indicated by the preceding derivative analysis. Firstly, samples that are accurately predicted with low uncertainty are reliably simple samples for the current model, while samples that are accurately predicted but have high uncertainty are samples within the learning scope of the model but do not yet have sufficient evidence to support the prediction. For the former, the regularization term in Eq.~\ref{tag:L} mitigates the influence of the KL divergence term, allowing the model to pay more attention to the accumulation of evidence on the target labels, while for the latter, it guides the correct growth of evidence by intensifying the inhibitory effect on the non-target label evidence.
Secondly, samples with incorrect predictions but high uncertainty indicate that the model makes the incorrect prediction due to the lack of sufficient evidence accumulation, while samples with incorrect predictions but low uncertainty suggest that the model makes a confident but incorrect prediction due to the accumulation of erroneous evidence. For the former, by reducing the coefficient in front of the KL divergence term, the model is allowed to explore more freely. For the latter, the regularization term increases its regulatory effect on imbalanced evidence by amplifying the coefficient in front of the KL divergence term, thereby suppressing the accumulation of evidence on incorrect class labels. Therefore, UASPL prioritizes reliably easy samples, selectively explores samples with insufficient evidence, and defers samples with misleading evidence, aligning with the easy-to-hard learning principle of SPL. The full procedure of UASPL is shown in Fig.~\ref{img:framework}, and the detailed algorithm of UASPL using the aforementioned loss function is presented in Algorithm~\ref{alg:algorithm1}.

\begin{algorithm}[!ht]
\caption{Uncertainty-Aware Self-Paced Learning (UASPL)}
\label{alg:algorithm1}
\KwIn{Training data $\{x_i, y_i\}_{i=1}^n$}, total epochs $EX$, inner epochs $I$
\KwOut{Trained classifier $f_\theta$}

Initialize model $f_\theta$, optimizer, number of classes $K$, the age parameter $\lambda$\

The initial model was pre-trained for 20 rounds using the traditional cross-entropy loss function

\For{$ex = 1$ \KwTo $EX$}{
    Compute Dirichlet parameters $\alpha_i$ and total evidence $S_i = \sum_{j=1}^K \alpha_{ij}$ for $i=1..n$}\;
    Compute uncertainty $u_i = K/S_i$, predicted probability $\mu_{ij} = \alpha_{ij}/S_i$, and correctness indicator $c_i = \mathbf{1}[\arg\max_j \mu_{ij} == y_i]$\;
    Compute coefficient $\text{coeff}_i = (1-c_i)(1-u_i) + c_i u_i$\;
    Compute sample-wise loss $\mathcal{L}_{\text{total}}^{(i)} = \mathcal{L}_{\text{EMSE}}^{(i)} + \text{coeff}_i \mathcal{L}_{\text{KL}}^{(i)}$\;
    Sort samples by $\mathcal{L}_{\text{total}}^{(i)}$ in ascending order\;
    Select a subset of samples $(x_s, y_s)$ with the smallest $\mathcal{L}_{\text{total}}^{(i)}$ as $\mathcal{D}_s$ based on $\lambda$\;

    \For{$inner = 1$ \KwTo $I$}{
        Compute $\alpha_i$, $S_i$, $u_i$, $\mu_{ij}$, $c_i$, $\text{coeff}_i$ for samples in $\mathcal{D}_s$}\;
        Compute the loss  $ \mathcal{L}_{\text{total}}^{(i)}$ for all selected samples\;
        Backpropagate and update model parameters $\theta$ \;
        \If{$\mathcal{L} < \epsilon$}{\textbf{break}}
    Update the age parameter $\lambda$
\end{algorithm}

\section{Experiments}
\label{sec_experiments}

In this section, comprehensive experiments are conducted to evaluate UASPL from multiple perspectives, including classification performance, interpretability, generality, and additional validation experiments. In addition, due to space limitations, part of the experimental results and analyses are provided in the Supplementary Material.

\subsection{Experimental Setting}
\label{subsec4.1}

This subsection describes the datasets, baselines and some additional details to facilitate understanding and reproduction.

\textbf{Dataset.} 
We evaluate our method on both tabular and image datasets. The tabular datasets include 25 datasets from the UCI repository, consisting of 10 balanced and 15 imbalanced datasets, with their basic information provided in Table A.1 of the Supplementary Material. The image datasets include four widely adopted classification benchmarks: CIFAR-10, FashionMNIST, MNIST, and SVHN.

\textbf{Baselines.} A total of fifteen baseline methods are included for comparison. These methods comprise conventional training (1), traditional SPL variants (2)--(4), SPL methods that incorporate additional information (5)--(8), and representative sample-reweighting methods (9)--(10). Moreover, 
as directly comparable SPL methods based on representative network-output-based and Bayesian uncertainty criteria are limited in the existing literature, we construct five adapted SPL baselines (11)--(15) by incorporating these uncertainty criteria into the SPL framework. The compared methods are briefly described as follows.

(1) \textit{Direct}: All samples are used directly for training without selection.

(2) \textit{SPL} \cite{kumar2010self}: Samples are selected based on the training loss in a traditional self-paced learning framework with hard regularization term.

(3) \textit{SPL\_linear} \cite{jiang2014easy}: Samples are selected based on the training loss in a traditional self-paced learning framework with linear regularization term.

(4) \textit{SPL\_mixture} \cite{zhao2015matrix}: Samples are selected based on the training loss in a traditional self-paced learning framework with mixture regularization term.

(5) \textit{CLU1} \cite{zhang2024weighted}: Samples are selected based on evidential uncertainty, where the basic belief assignment (BBA) is modeled by EKNN.

(6) \textit{CLU2} \cite{zhang2024weighted}: Samples are selected based on evidential uncertainty, with BBA modeled by ECM.

(7) \textit{WSPLBF1} \cite{zhang2024weighted}: Sample selection is guided by both evidential uncertainty and training loss, with BBA modeled by EKNN.

(8) \textit{WSPLBF2} \cite{zhang2024weighted}: Sample selection is guided by both evidential uncertainty and training loss, with BBA modeled by ECM.

(9) \textit{MW-Net} \cite{shu2019meta}: A representative sample-reweighting method that adaptively learns an explicit weighting function directly from unbiased meta-data.

(10) \textit{active\_bias} \cite{chang2017active}: A representative sample-weighting method that emphasizes samples with high predictive variance during training.

(11) \textit{SPL\_conf} \cite{islam2023paced}: An adapted SPL baseline that uses confidence scores from the model for sample selection.

(12) \textit{SPL\_marg} \cite{feng2025your}: An adapted SPL baseline that uses margin-based difficulty scores from the model for sample selection.

(13) \textit{MC\_MI} \cite{gal2016dropout,gal2017deep}: An adapted SPL baseline that uses mutual information estimated via MC dropout as an uncertainty score for sample selection.

(14) \textit{MC\_pBALD} \cite{gal2016dropout,kim2021task}: An adapted SPL baseline that uses power-scaled mutual information estimated via MC dropout as an uncertainty score for sample selection.

(15) \textit{MC\_BalEnt} \cite{gal2016dropout,woo2023active}: An adapted SPL baseline that uses balanced entropy estimated via MC dropout as an uncertainty score for sample selection.

\textbf{Additional Details.}
All methods adopt the same multilayer perceptron backbone on the UCI datasets, which is pretrained for 20 epochs using cross-entropy loss \cite{li2022tedl}. Regarding data partitioning, 50\% of the samples in each dataset are randomly assigned to the test set. For all methods except \textit{MW-Net}, the remaining 50\% serve as the training set. For \textit{MW-Net}, the same test set is retained, while the remaining half is further split into training and meta-data sets at a 4:1 ratio, resulting in an overall split of 50\% test set, 40\% training set, and 10\% meta-data set \cite{shu2019meta, ShuPmw, ShuCMW}. To mitigate the impact of random partitioning, 50 Monte Carlo runs are performed on each dataset, and the average results are reported. Furthermore, \textit{CLU1}, \textit{CLU2}, \textit{WSPLBF1}, and \textit{WSPLBF2} \cite{zhang2024weighted} consider both sample difficulty and class proportion during sample selection, i.e., samples are selected in the same proportion for each class, whereas the other methods focus solely on sample difficulty. For SPL-related methods, the training process is divided into six stages: the first stage uses 25\% of the training samples, and each subsequent stage adds an additional 15\%.

\subsection{Classification Performance}
\label{subsec4.2}

Table~\ref{tab:uci_summary_all} presents the overall accuracy comparisons in terms of both Accuracy Mean and Accuracy Std. For each metric, the average value, average rank, and best count across datasets are reported. Detailed accuracy results on datasets are provided in the Supplementary Material (Tables B.2 and B.3). The overall comparisons for F1-score, Precision, and Recall are given in the Supplementary Material (Tables B.4--B.12), while the main numerical results are summarized below.

\begin{table}[!ht]
\centering
\caption{Comparison of all methods on datasets in terms of accuracy. }
\label{tab:uci_summary_all}
\fontsize{10pt}{12pt}\selectfont
\renewcommand{\arraystretch}{1.12} 
\setlength{\tabcolsep}{2.2pt} 
\begin{tabular}{>{\centering\arraybackslash}p{0.13\linewidth}cccccc}
\toprule
\multirow{2}{*}{Method} & \multicolumn{3}{c}{Accuracy Mean} & \multicolumn{3}{c}{Accuracy Std.} \\
\cmidrule(lr){2-4} \cmidrule(lr){5-7}
& Avg. Value & Avg. Rank & Best Count & Avg. Value & Avg. Rank & Best Count \\
\midrule
Direct & 0.7986 & 5.88 & 0 & 0.0461 & 7.90 & 2 \\
SPL & 0.8143 & 4.60 & 0 & 0.0405 & 8.08 & 0 \\
SPL\_linear & \underline{0.8255} & \underline{3.16} & 4 & \underline{0.0296} & 7.24 & 1 \\
SPL\_mixture & 0.8179 & 4.60 & 0 & 0.0445 & 8.48 & 1 \\
CLU1 & 0.7961 & 6.96 & 0 & 0.0555 & 8.82 & 1 \\
CLU2 & 0.7862 & 6.48 & 0 & 0.0606 & 9.48 & 1 \\
WSPLBF1 & 0.8087 & 5.28 & 1 & 0.0522 & 8.72 & 1 \\
WSPLBF2 & 0.8041 & 5.24 & 0 & 0.0455 & 8.80 & 2 \\
MW-Net & 0.8123 & 4.96 & 1 & 0.0304 & 6.94 & 1 \\
active\_bias & 0.7775 & 7.88 & 0 & 0.0586 & 12.60 & 0 \\
SPL\_conf & 0.7791 & 7.48 & 1 & 0.0754 & 8.54 & 2 \\
SPL\_marg & 0.7808 & 6.92 & 1 & 0.0737 & 9.14 & 0 \\
MC\_MI & 0.7527 & 8.28 & 1 & 0.0739 & 9.40 & 2 \\
MC\_pBALD & 0.7444 & 8.44 & 2 & 0.0727 & 9.64 & 4 \\
MC\_BalEnt & 0.7432 & 8.12 & \underline{6} & 0.0621 & \underline{6.60} & \underline{5} \\
UASPL & \textbf{0.8285} & \textbf{1.76} & \textbf{13} & \textbf{0.0271} & \textbf{5.62} & \textbf{8} \\
\bottomrule
\end{tabular}

\par 
\vspace{5pt}
\begin{minipage}{\textwidth}
\raggedright
\footnotesize
\textit{Note:} For each dataset, “Accuracy Mean” and “Accuracy Std.” are computed from 50 independent random trials. The sub-items under each metric denote: (i) the average value of this metric across datasets, (ii) the average rank of this metric across datasets, and (iii) the win count (number of datasets) where the method achieves the best value of this metric. Bold and underlined values indicate the best and second-best overall results, respectively.
\end{minipage}

\end{table}

In terms of accuracy, as shown in Table~\ref{tab:uci_summary_all}, UASPL achieves the highest average accuracy (0.8285), the best average rank (1.76), and the largest number of wins (13 out of 25 datasets). 
Its average standard deviation (0.0271) is also the smallest among all methods, indicating superior stability. 
For F1-score (Tables B.4--B.6 in the Supplementary Material), UASPL ties for the highest mean value (0.7720), while attaining the best average rank (4.88) and the most wins (12). 
In terms of Precision (Tables B.7--B.9), UASPL ranks first in mean value (0.7884), average rank (3.76), and win count (14). 
For Recall (Tables B.10--B.12), UASPL again leads in mean value (0.7760), average rank (4.52), and win count (13), and also achieves the smallest average standard deviation (0.0336). In summary, UASPL achieves the strongest performance in terms of accuracy, with the highest mean accuracy and the lowest average standard deviation across the reported summary statistics. Moreover, it also obtains highly competitive results on F1-score, Precision, and Recall, which further demonstrates the effectiveness of UASPL.

\begin{figure}[!ht]
  \centering
\begin{subfigure}[b]{0.48\textwidth} 
  \centering
  \includegraphics[width=\linewidth]{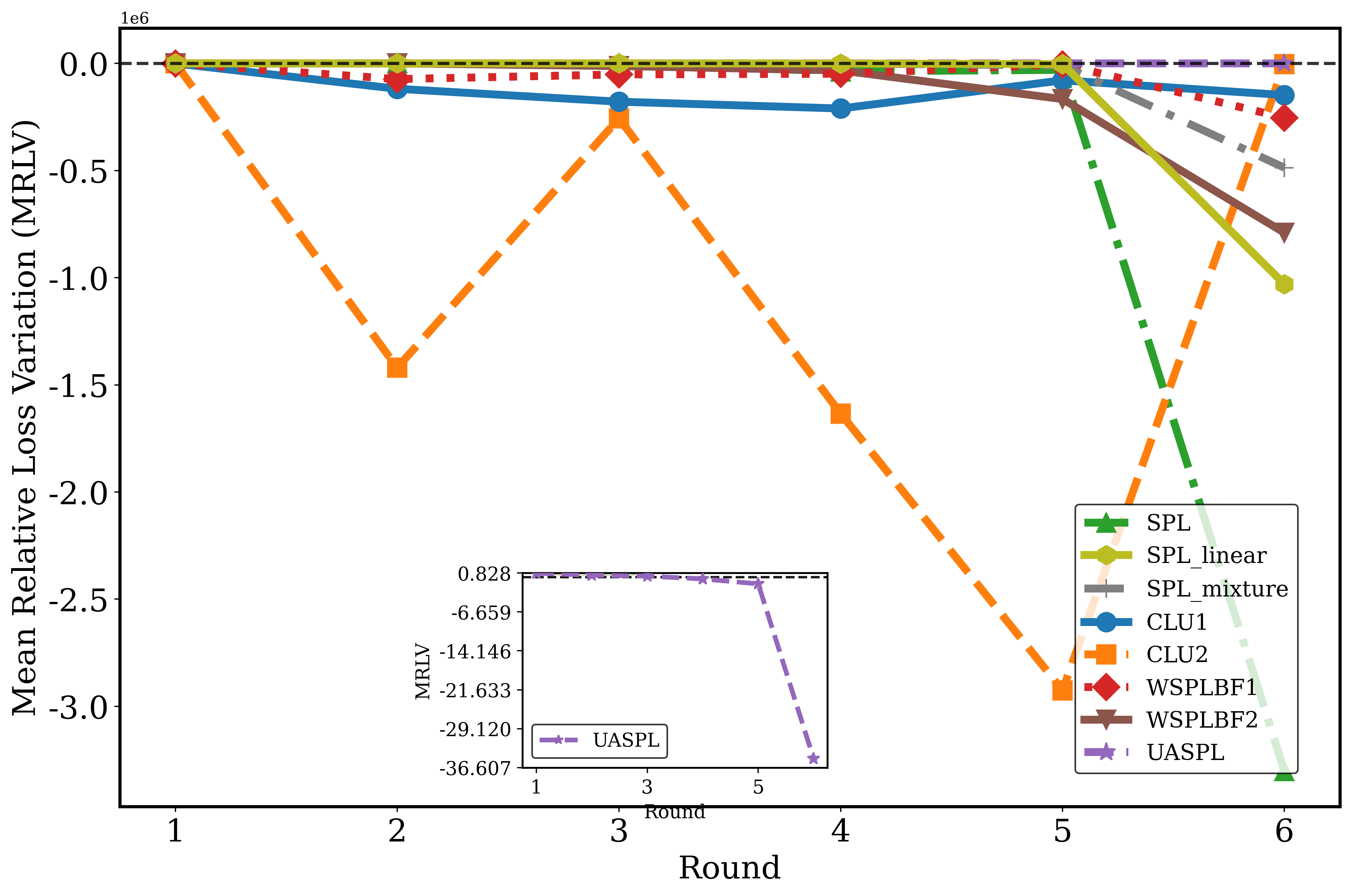}
  \caption{Transfusion}
  
\end{subfigure}
\hfill
\begin{subfigure}[b]{0.48\textwidth}
  \centering
  \includegraphics[width=\linewidth]{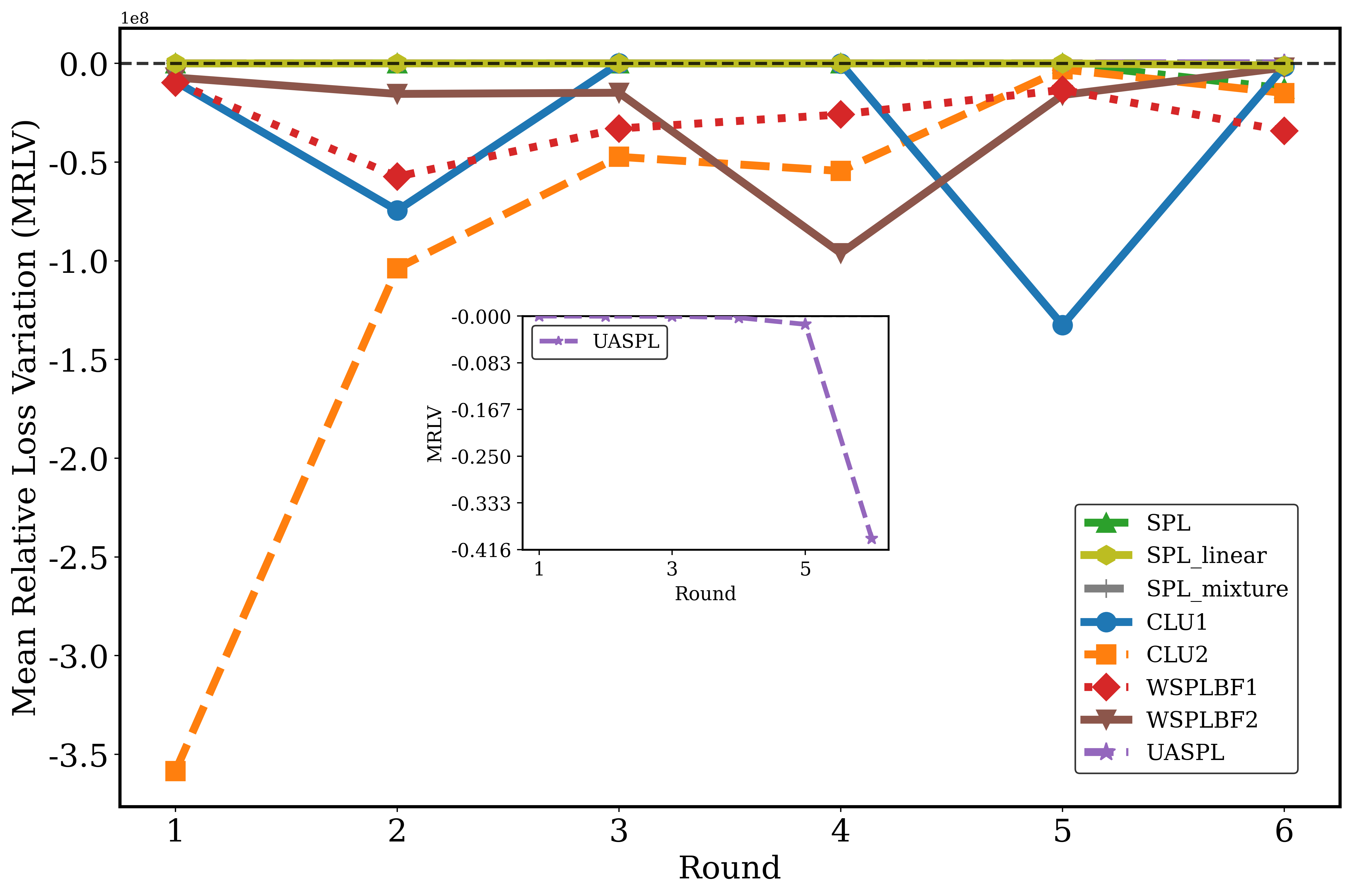}
  \caption{Wine}
  
\end{subfigure}
\vspace{0.5em}
\begin{subfigure}[b]{0.48\textwidth}  
  \centering
  \includegraphics[width=\linewidth]{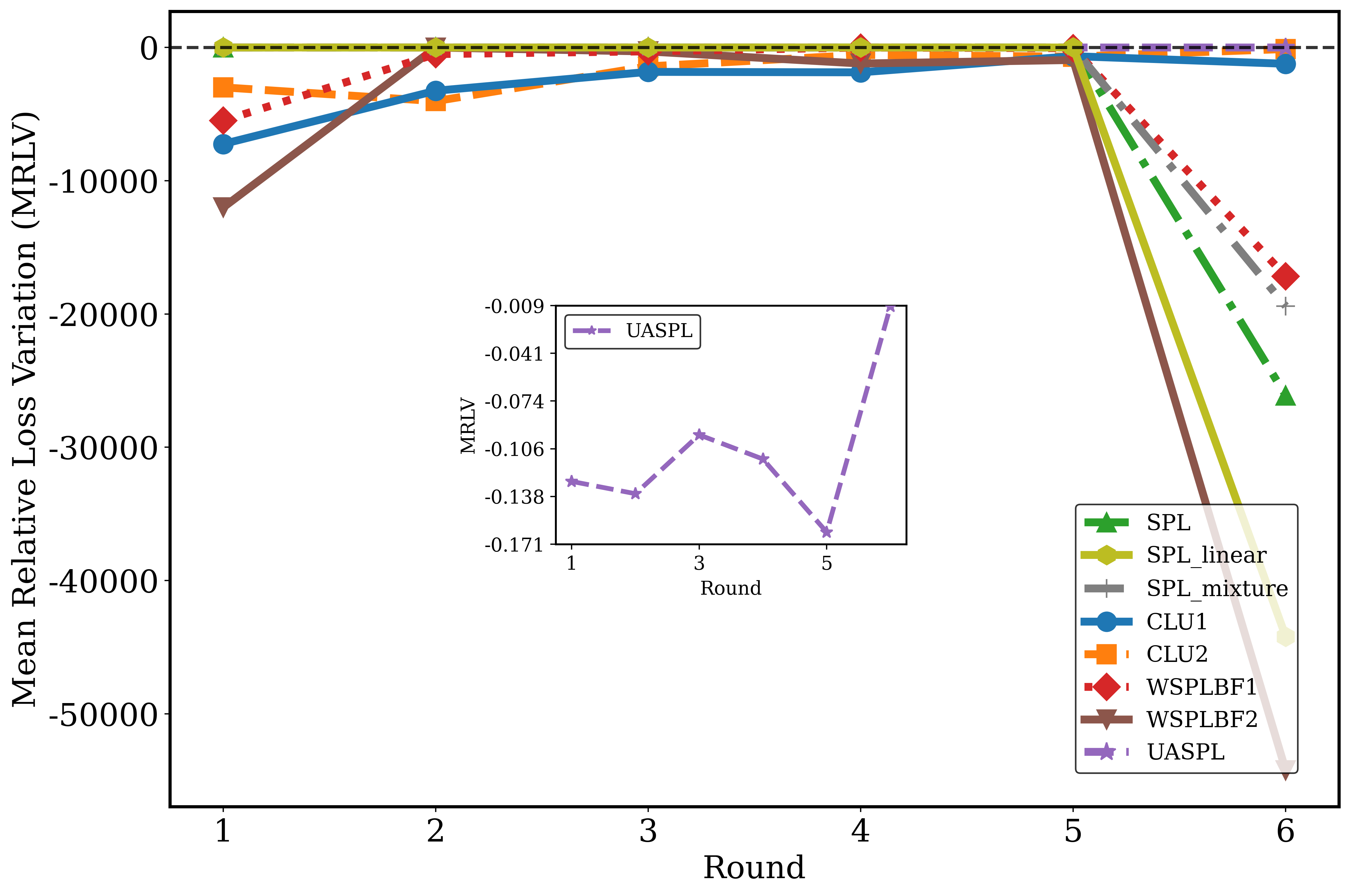}
  \caption{Pop}
  
\end{subfigure}
\hfill
\begin{subfigure}[b]{0.48\textwidth}  
  \centering
  \includegraphics[width=\linewidth]{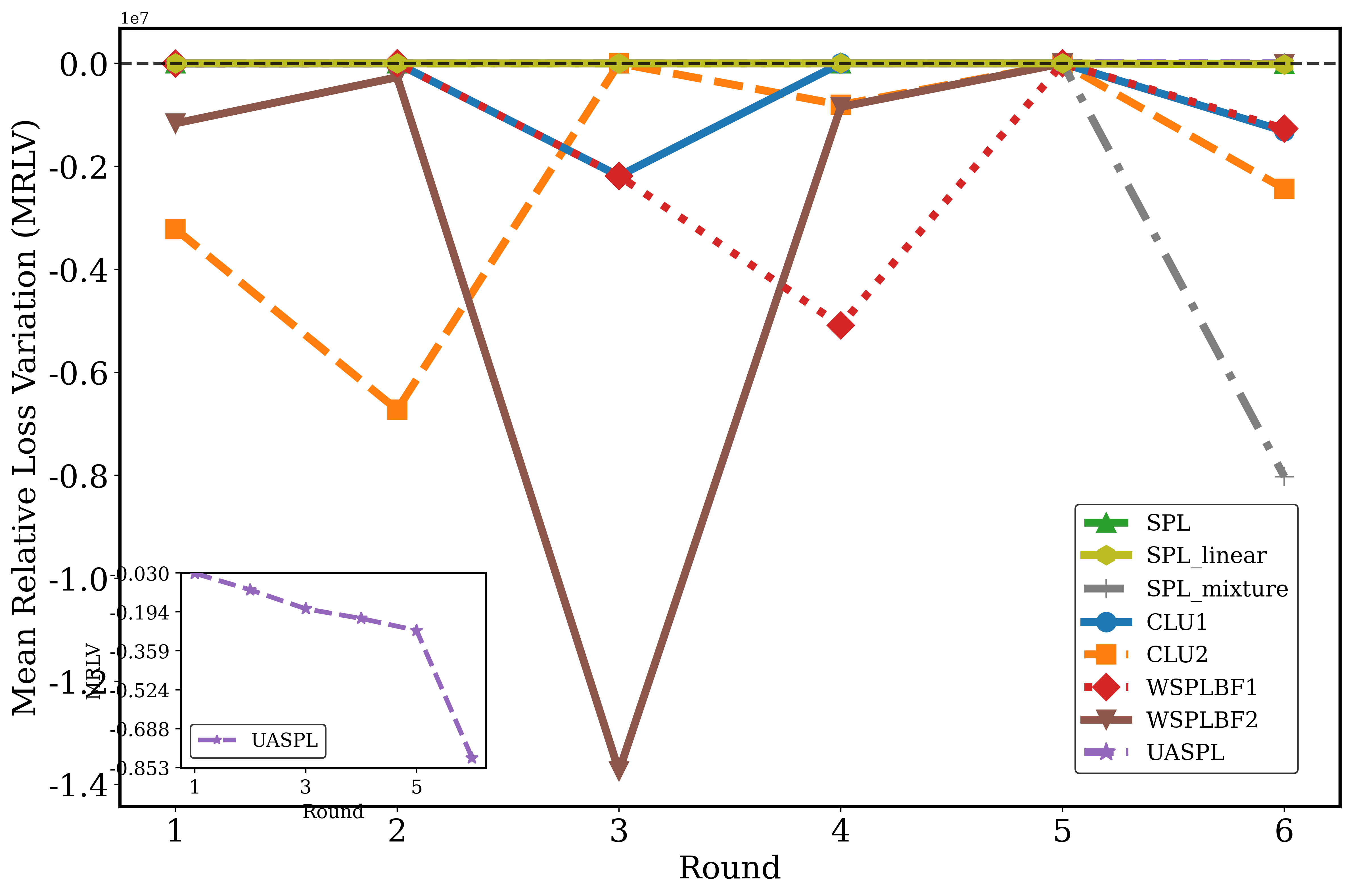}
  \caption{Breast\_Cancer\_Diagnostic}
  
\end{subfigure}
\vspace{0.5em}
\begin{subfigure}[b]{0.48\textwidth}  
  \centering
  \includegraphics[width=\linewidth]{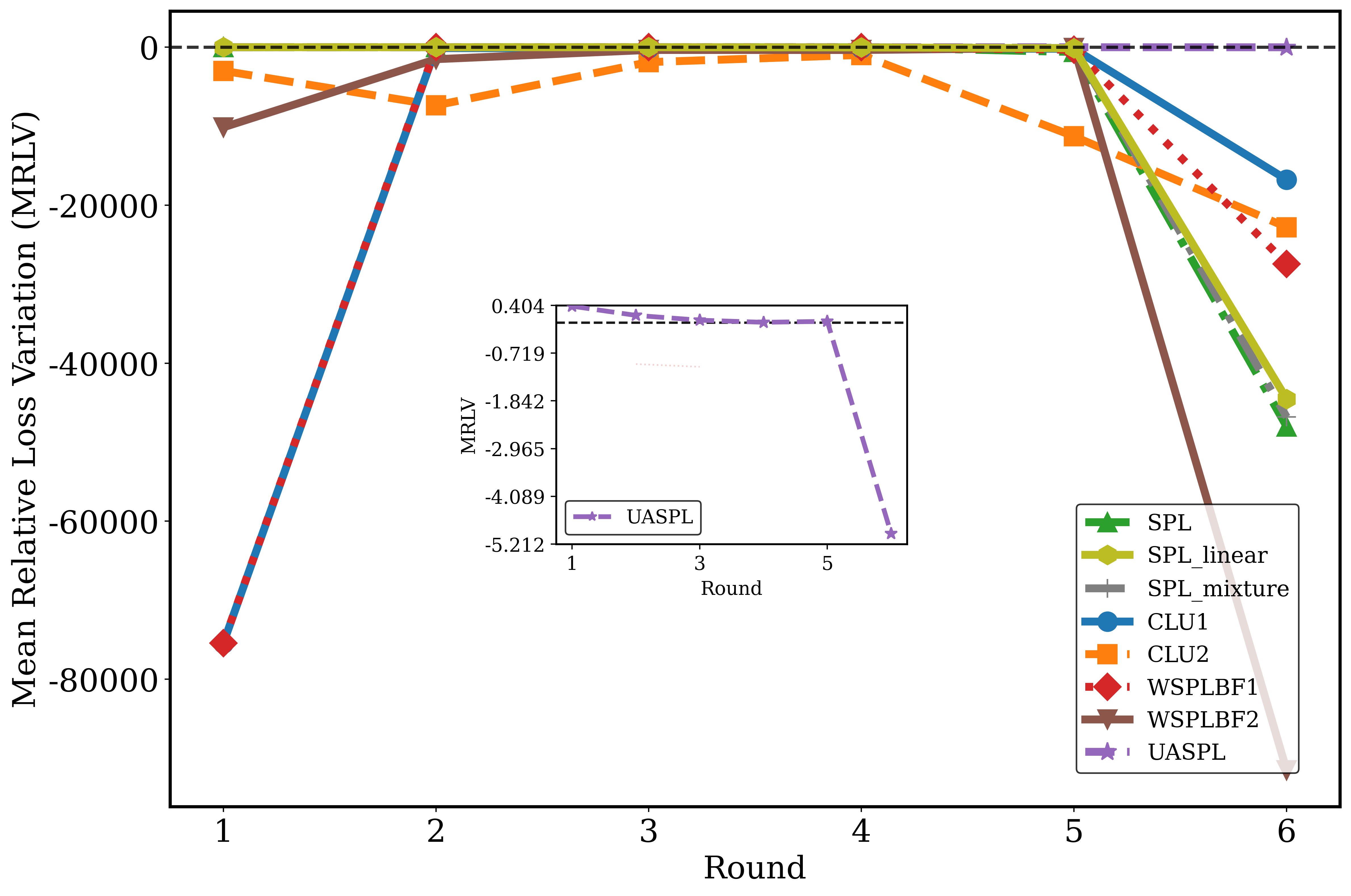}
  \caption{Heart}
  
\end{subfigure}
\hfill
\begin{subfigure}[b]{0.48\textwidth}  
  \centering
  \includegraphics[width=\linewidth]{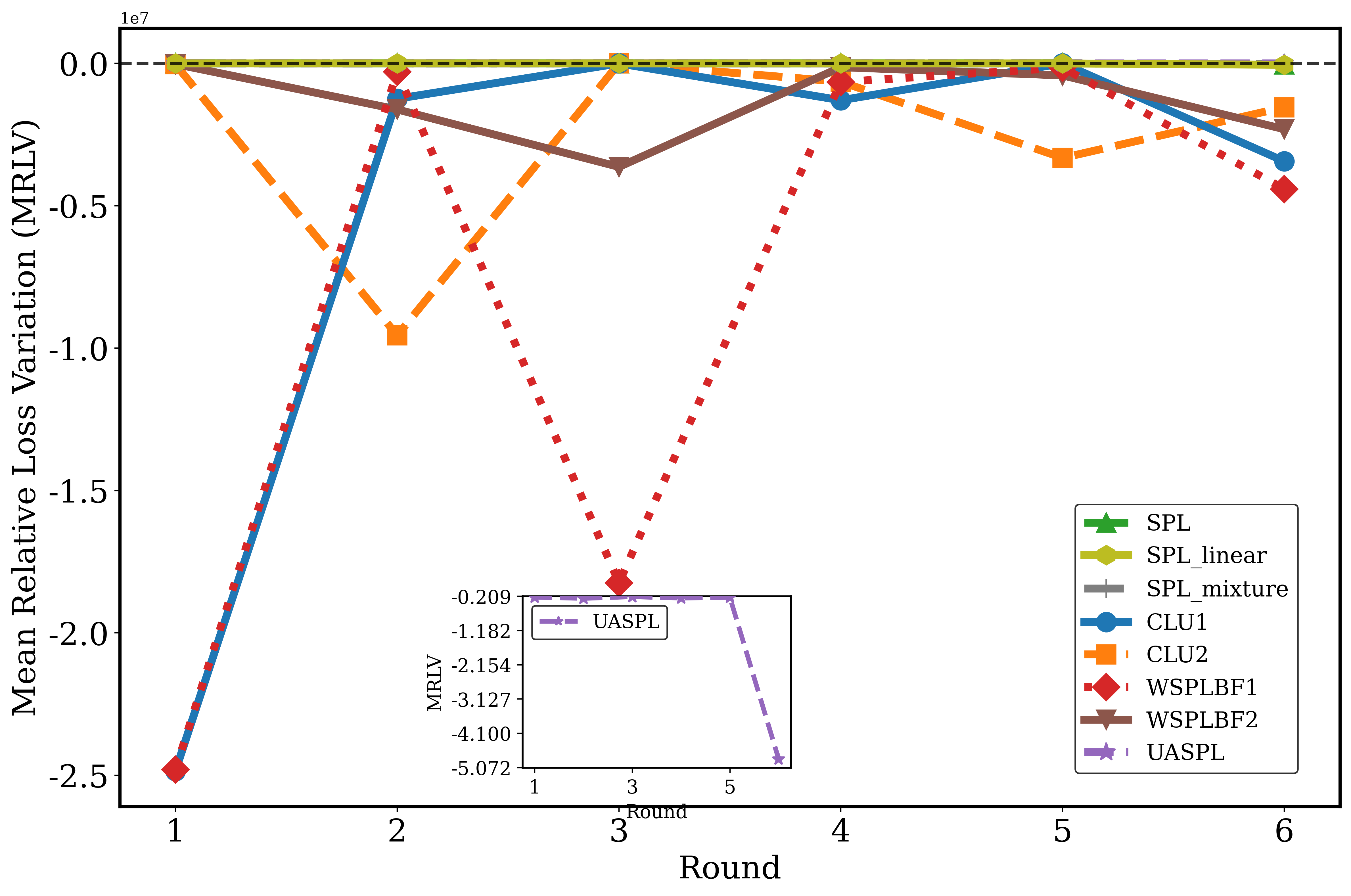}
  \caption{Ionosphere}
  
\end{subfigure}
\caption{Comparison of mean relative loss variation across rounds for UASPL and representative baselines on six datasets. For readability, the corresponding comparison with Baselines 11--15 is reported separately in Figure B.1 of the Supplementary Material.}
\label{fig:MRLV_UASPL}
\end{figure}

Additionally, as illustrated in the Fig.~\ref{fig:MRLV_UASPL}, the mean relative loss variation of representative baselines fluctuates drastically, with nearly all values being negative, while UASPL exhibits a slightly overall declining trend with only minor occasional fluctuations, and can achieve positive values in the initial rounds across several datasets. This indicates that before the introduction of highly challenging samples, the model's learning for the samples selected by UASPL in the first round is relatively stable (the loss does not exhibit significant fluctuations), indicating that compared with other methods, the simple samples selected by UASPL are more conducive to the model’s learning in subsequent stages.

\subsection{Interpretability Analysis}
\label{subsec4.3}

Given the theoretical principle of UASPL provided in Subsection \ref{subsec3.2}, a further analysis is conducted on the interpretability of UASPL in this subsection. 

\begin{figure}[!ht]
  \centering

\begin{subfigure}[b]{0.48\textwidth}  
  \includegraphics[width=\linewidth]{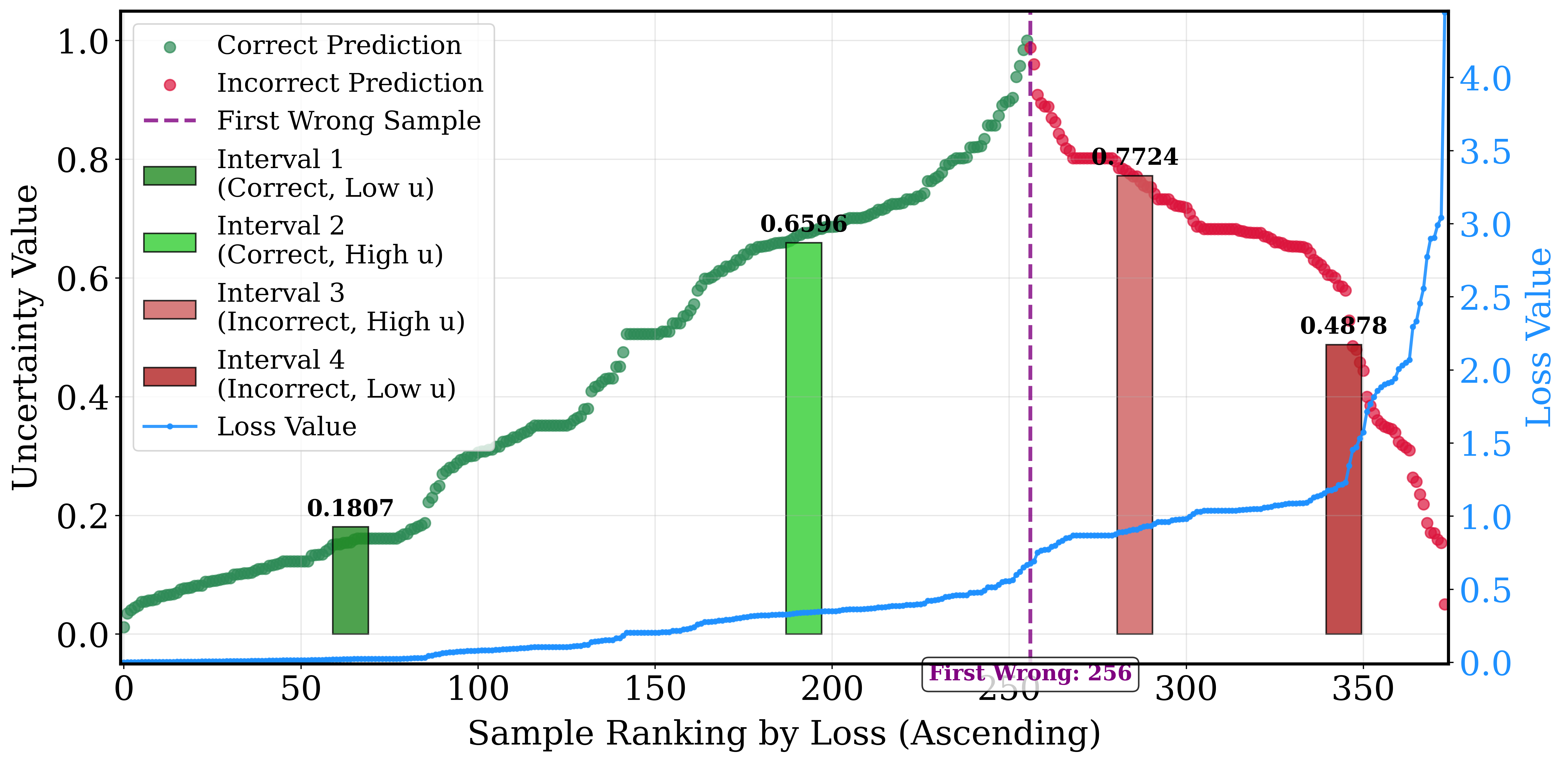}
  \caption{Round 1}
\end{subfigure}
\hfill
\begin{subfigure}[b]{0.48\textwidth}
  \includegraphics[width=\linewidth]{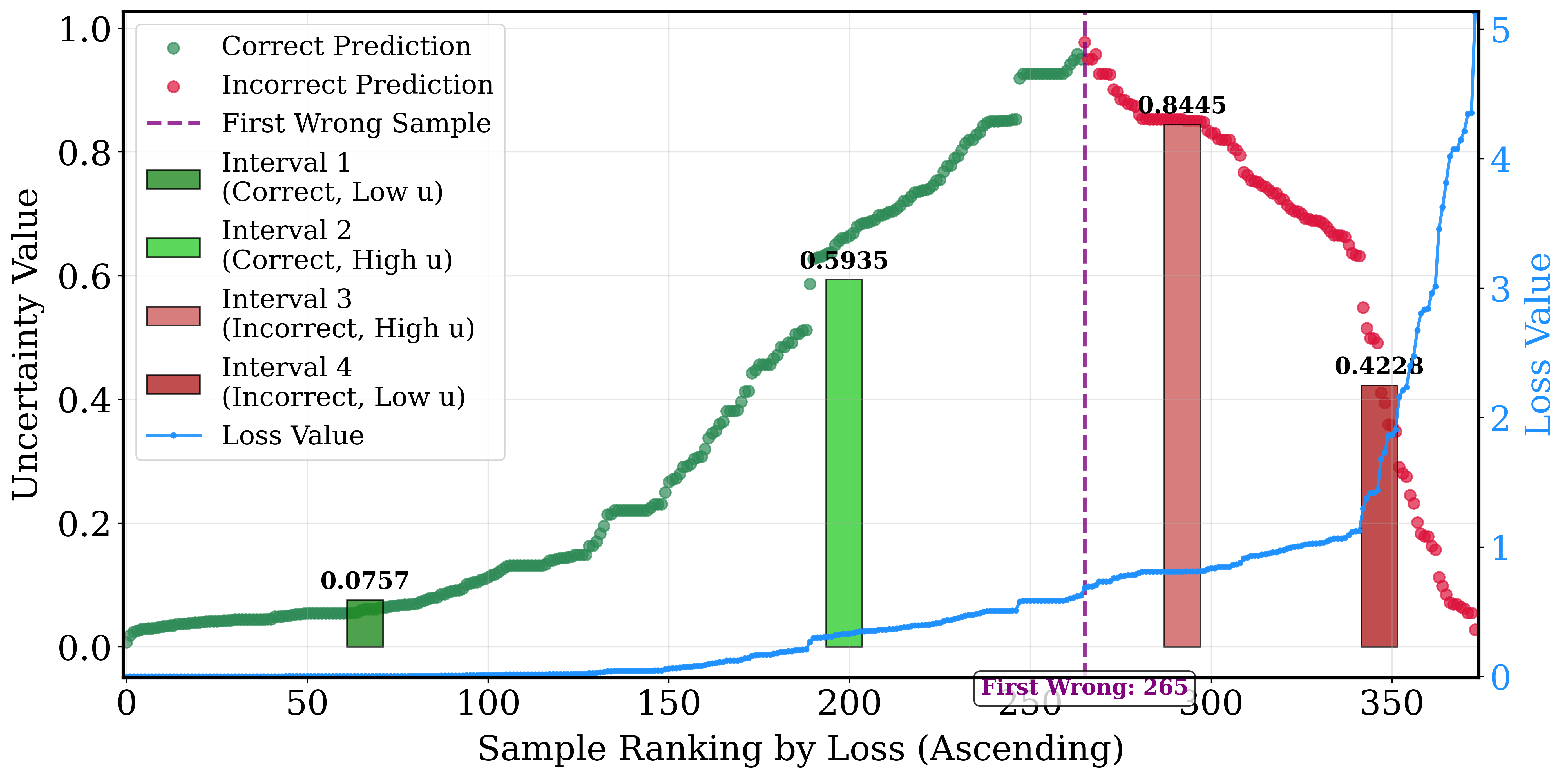}
  \caption{Round 2}
\end{subfigure}

\vspace{0.5em}

\begin{subfigure}[b]{0.48\textwidth}  
  \includegraphics[width=\linewidth]{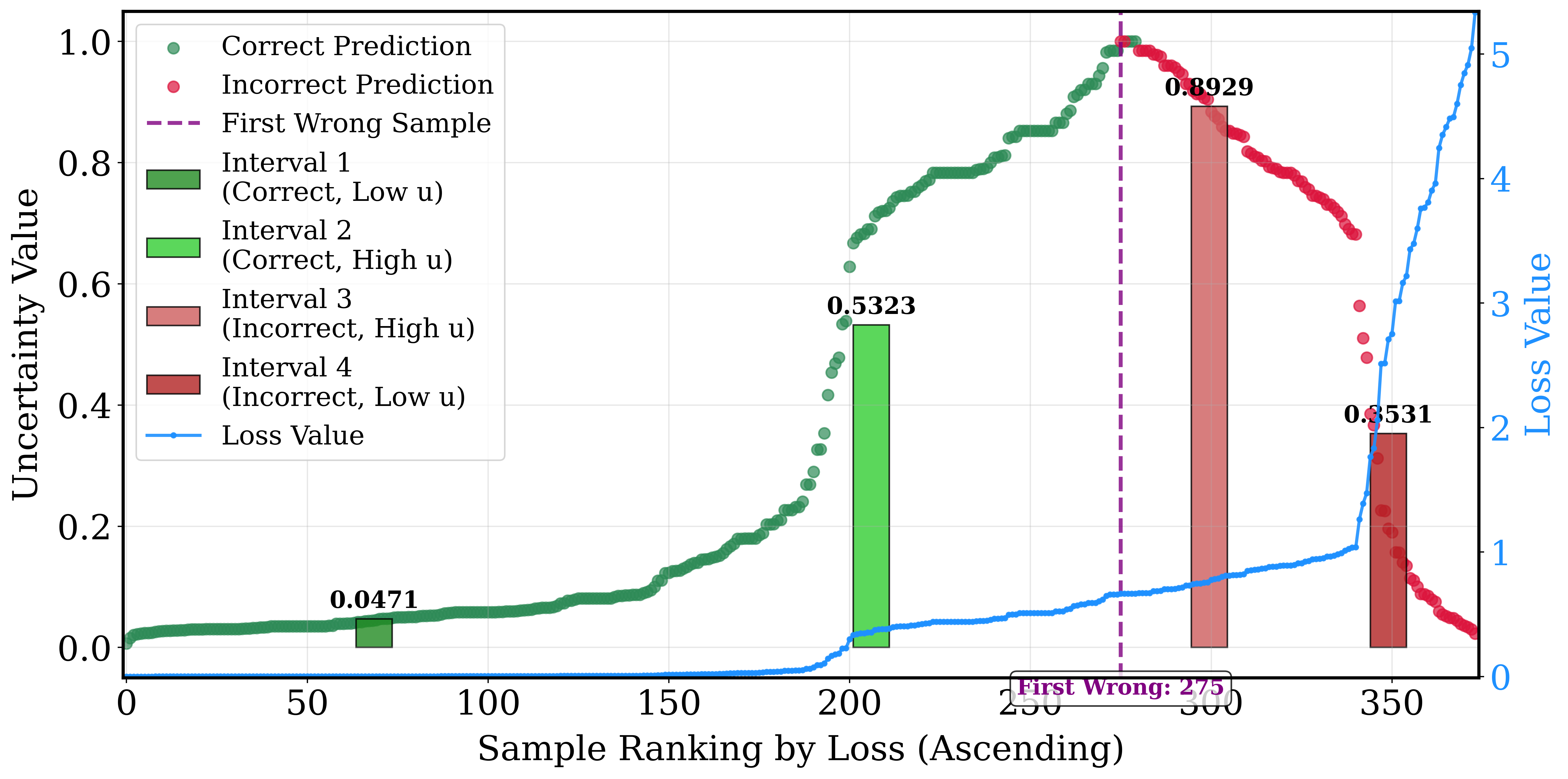}
  \caption{Round 3}
\end{subfigure}
\hfill
\begin{subfigure}[b]{0.48\textwidth}  
  \includegraphics[width=\linewidth]{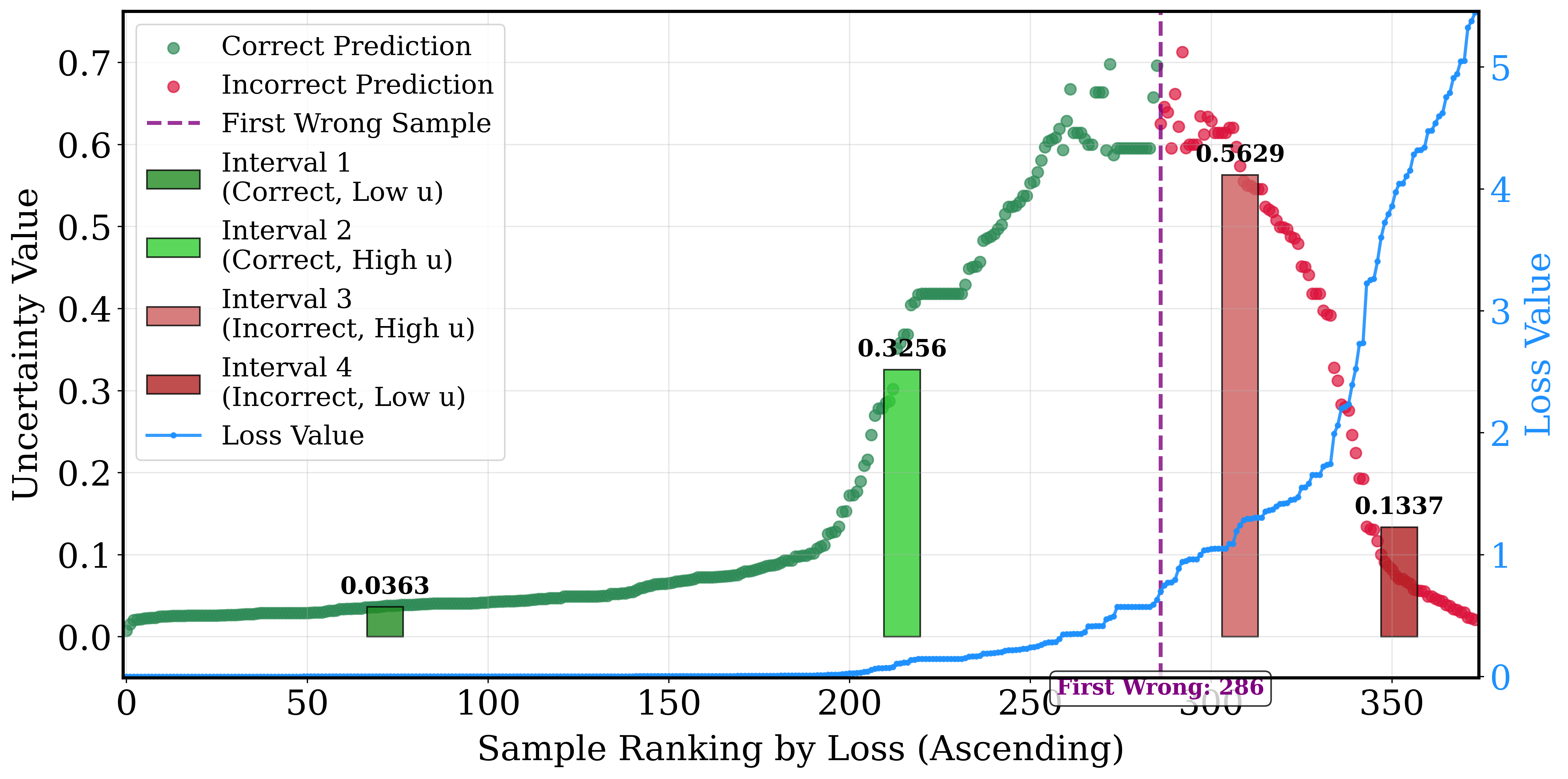}
  \caption{Round 4}
\end{subfigure}

\vspace{0.5em}

\begin{subfigure}[b]{0.48\textwidth}  
  \includegraphics[width=\linewidth]{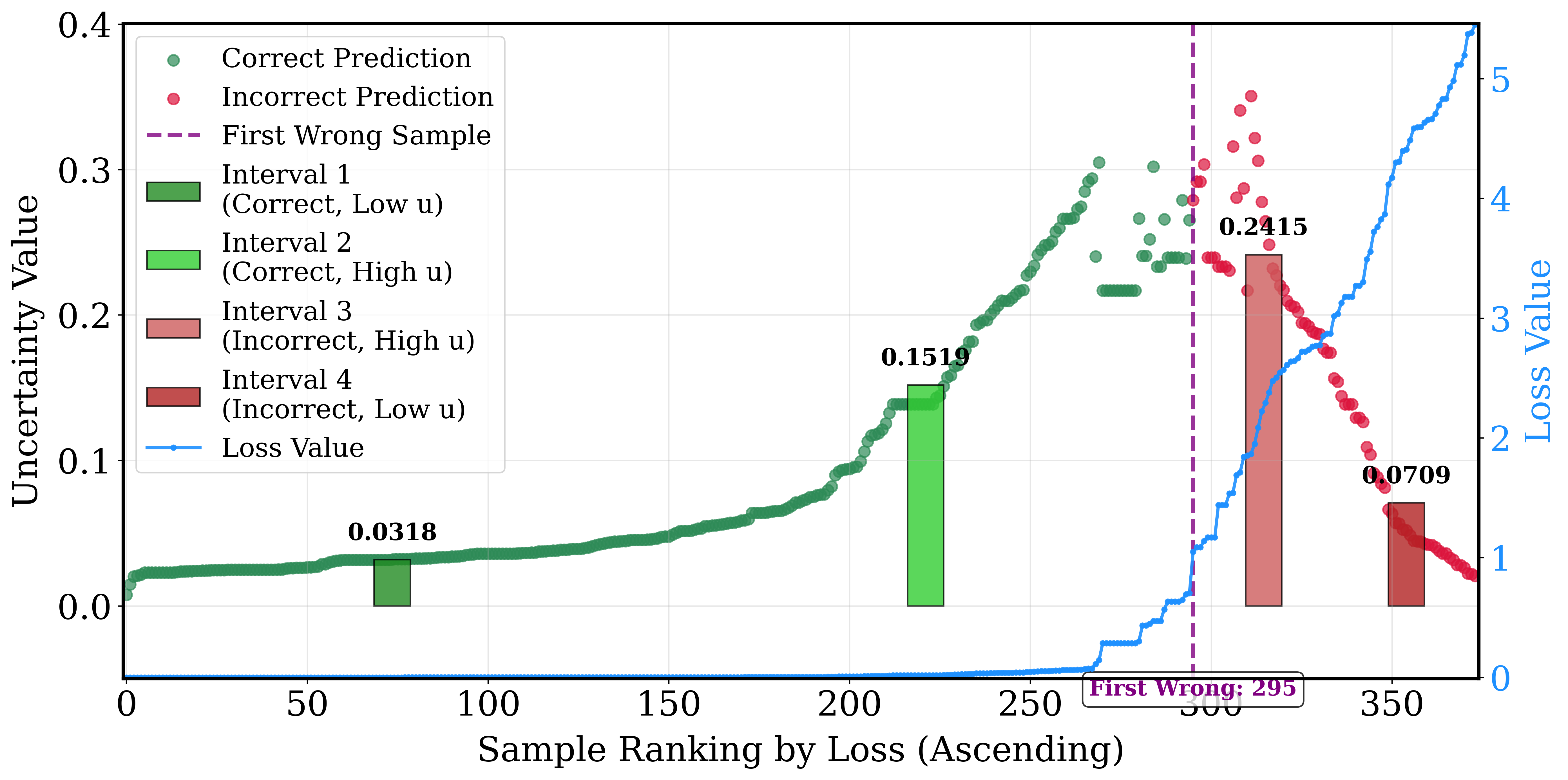}
  \caption{Round 5}
\end{subfigure}
\hfill
\begin{subfigure}[b]{0.48\textwidth} 
  \includegraphics[width=\linewidth]{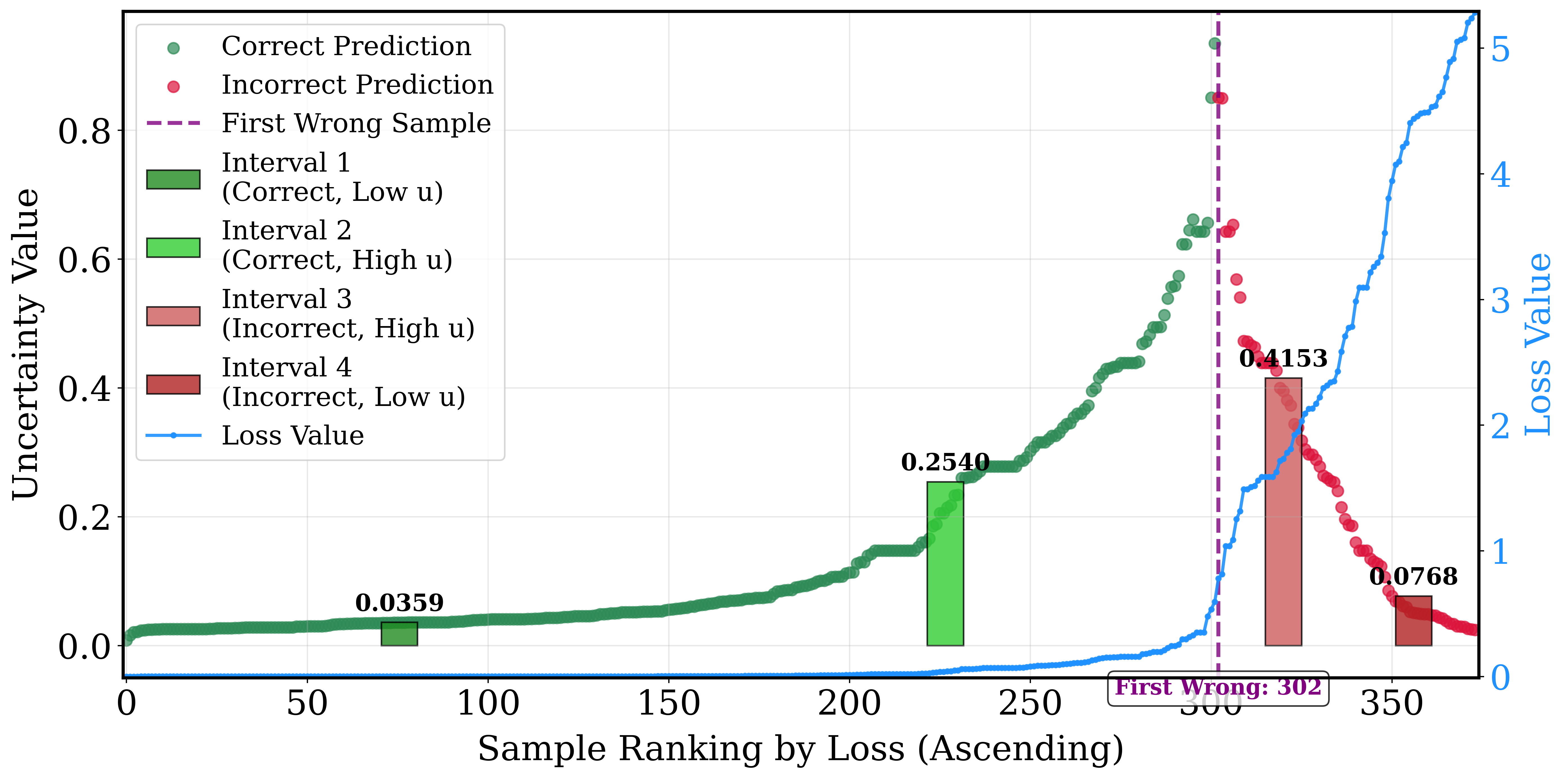}
  \caption{Round 6}
\end{subfigure}
 \caption{Prediction results and uncertainty of UASPL with loss ascending sorting on the Transfusion dataset.}
  \label{fig:analysis_UASPL}
\end{figure}

To analyze the sample selection strategy of UASPL, Fig.~\ref{fig:analysis_UASPL} visualizes the characteristics of the sample distribution across six rounds of sample selection, where samples are ranked by ascending loss values calculated via Eq.~\ref{tag:L_total_def}. The scatter plot in the figure marks correctly predicted samples in green and incorrectly predicted ones in red, clearly reflecting that among samples ordered by ascending loss, correctly predicted samples are prioritized over incorrectly predicted ones in the sample selection of UASPL. The four colored bars correspond respectively to the four progressively increasing difficulty intervals mentioned in Subsection \ref{subsec3.2} regarding the sample selection preference: all samples ranked by ascending loss are split into two major parts according to the index of the first incorrectly predicted sample, and each part is then bisected again at its median sample, thereby forming four intervals. The bold numerical labels on these bars display the mean uncertainty of each interval.
Obviously, for correctly predicted samples, those with lower uncertainty are prioritized over those with higher uncertainty. In contrast, for incorrectly predicted samples, those with higher uncertainty are granted higher priority than their lower-uncertainty counterparts. This sample selection preference enables a gradual transition from reliably simple samples to complex and challenging samples. Therefore, the sample selection of UASPL is highly clear and interpretable.

\subsection{Generality Evaluation}
\label{subsec4.4}

In the aforementioned experiments, UASPL employs the hard regularizer for comparison with the baseline methods. To evaluate the generality of UASPL across different self-paced learning variants, this subsection replaces the hard regularizer in UASPL with the linear regularizer \cite{jiang2014easy} and the mixture regularizer \cite{zhao2015matrix}, yielding two modified variants denoted as \textit{UASPL\_linear} and \textit{UASPL\_mixture}, respectively. Due to space limitations, additional generality evaluation results, including the illustrative analysis of training behavior and sample-selection characteristics for \textit{UASPL\_linear} and \textit{UASPL\_mixture}, together with the detailed classification results on datasets, are provided in Appendix C of the Supplementary Material.

\begin{table}[!ht]
\centering
\caption{Comparison of all methods on datasets in terms of Accuracy for the UASPL\_linear generalization experiment.}
\label{tab:gen_linear_accuracy_summary}
\small
\renewcommand{\arraystretch}{1.12}
\setlength{\tabcolsep}{3pt}
\begin{tabular}{>{\centering\arraybackslash}p{0.20\linewidth}cccccc}
\toprule
\multirow{2}{*}{Method} & \multicolumn{3}{c}{Accuracy Mean} & \multicolumn{3}{c}{Accuracy Std.} \\
\cmidrule(lr){2-4} \cmidrule(lr){5-7}
& Avg. Value & Avg. Rank & Best Count & Avg. Value & Avg. Rank & Best Count \\
\midrule
Direct & 0.7986 & 9.72 & 0 & 0.0461 & 7.28 & 2 \\
SPL & 0.8143 & 6.76 & 1 & 0.0405 & 7.44 & 0 \\
SPL\_linear & \underline{0.8255} & \underline{4.36} & 1 & \textbf{0.0296} & 6.64 & 1 \\
CLU1 & 0.7961 & 9.56 & 0 & 0.0555 & 7.96 & 2 \\
CLU2 & 0.7862 & 10.24 & 0 & 0.0606 & 8.76 & 1 \\
WSPLBF1 & 0.8087 & 7.40 & 0 & 0.0522 & 8.12 & 1 \\
WSPLBF2 & 0.8041 & 7.72 & 1 & 0.0455 & 8.16 & 3 \\
MW-Net & 0.8123 & 7.12 & 1 & \underline{0.0304} & 6.40 & 2 \\
active\_bias & 0.7775 & 10.84 & 0 & 0.0586 & 11.68 & 0 \\
SPL\_conf & 0.7791 & 9.72 & 1 & 0.0754 & 7.92 & 2 \\
SPL\_marg & 0.7808 & 9.40 & 1 & 0.0737 & 8.44 & 0 \\
MC\_MI & 0.7527 & 7.36 & 1 & 0.0739 & 8.68 & 2 \\
MC\_pBALD & 0.7444 & 8.36 & 2 & 0.0727 & 8.88 & \underline{4} \\
MC\_BalEnt & 0.7432 & 5.96 & \underline{7} & 0.0621 & \underline{6.04} & \textbf{5} \\
UASPL\_linear & \textbf{0.8284} & \textbf{4.28} & \textbf{13} & 0.0311 & \textbf{5.68} & \underline{4} \\
\bottomrule
\end{tabular}
\end{table}

\begin{table}[!ht]
\centering
\caption{Comparison of all methods on datasets in terms of Accuracy for the UASPL\_mixture generalization experiment.}
\label{tab:gen_mixture_accuracy_summary}
\small
\renewcommand{\arraystretch}{1.12}
\setlength{\tabcolsep}{3pt}
\begin{tabular}{>{\centering\arraybackslash}p{0.20\linewidth}cccccc}
\toprule
\multirow{2}{*}{Method} & \multicolumn{3}{c}{Accuracy Mean} & \multicolumn{3}{c}{Accuracy Std.} \\
\cmidrule(lr){2-4} \cmidrule(lr){5-7}
& Avg. Value & Avg. Rank & Best Count & Avg. Value & Avg. Rank & Best Count \\
\midrule
Direct & 0.7986 & 9.56 & 0 & 0.0461 & 7.12 & 2 \\
SPL & 0.8143 & 6.52 & 1 & 0.0405 & 7.32 & 0 \\
SPL\_mixture & \underline{0.8179} & \underline{5.76} & 0 & 0.0445 & 7.84 & 0 \\
CLU1 & 0.7961 & 9.40 & 0 & 0.0555 & 7.88 & 2 \\
CLU2 & 0.7862 & 10.24 & 0 & 0.0606 & 8.64 & 1 \\
WSPLBF1 & 0.8087 & 7.00 & 1 & 0.0522 & 7.92 & 1 \\
WSPLBF2 & 0.8041 & 7.56 & 2 & 0.0455 & 8.00 & 3 \\
MW-Net & 0.8123 & 6.88 & 2 & \underline{0.0304} & 6.16 & 2 \\
active\_bias & 0.7775 & 10.68 & 0 & 0.0586 & 11.60 & 0 \\
SPL\_conf & 0.7791 & 9.76 & 1 & 0.0754 & 7.88 & 1 \\
SPL\_marg & 0.7808 & 9.44 & 1 & 0.0737 & 8.32 & 0 \\
MC\_MI & 0.7527 & 7.28 & 1 & 0.0739 & 8.64 & 2 \\
MC\_pBALD & 0.7444 & 8.32 & 2 & 0.0727 & 8.96 & 3 \\
MC\_BalEnt & 0.7432 & 6.00 & \underline{6} & 0.0621 & \underline{6.08} & \underline{5} \\
UASPL\_mixture & \textbf{0.8285} & \textbf{4.40} & \textbf{12} & \textbf{0.0290} & \textbf{5.92} & \textbf{7} \\
\bottomrule
\end{tabular}
\end{table}

The corresponding classification results on the datasets are reported in Table~\ref{tab:gen_linear_accuracy_summary} (linear regularizer) and Table~\ref{tab:gen_mixture_accuracy_summary} (mixture regularizer). As shown in both tables, \textit{UASPL\_linear} and \textit{UASPL\_mixture} achieve the highest average accuracy (0.8284 and 0.8285, respectively) and the best (lowest) average rank (4.28 and 4.40) among all compared methods. In terms of best count (the number of datasets on which a method achieves the highest accuracy), \textit{UASPL\_linear} obtains 13 best results (the highest), and \textit{UASPL\_mixture} obtains 12 best results (also the highest). For the standard deviation of accuracy, \textit{UASPL\_mixture} achieves the lowest average value (0.0290) and the best average rank (5.92), while \textit{UASPL\_linear} ranks second in both metrics. These results demonstrate that the proposed uncertainty-aware sample selection mechanism remains effective under different self-paced regularization forms (hard, linear, and mixture), confirming the strong generality of UASPL across various SPL variants.

\subsection{Additional Experimental Validation}
\label{subsec4.5}

To further validate the effectiveness of UASPL, we conduct additional experiments covering five aspects: ablation studies, robustness to label noise, sensitivity to pretraining, statistical significance testing, and evaluation on image benchmarks.

Due to page limitations, the main text focuses on the ablation studies and the evaluation on image benchmarks. The other three experiments are reported in detail in the Supplementary Material and are briefly summarized here. Specifically, Appendix~E presents the robustness experiments on the datasets, showing that UASPL remains competitive under both uniform and asymmetric label noise, with UASPL\_linear exhibiting particularly stable performance under severe corruption. Appendix~F reports the pretraining sensitivity analysis, indicating that the adopted pretraining duration lies within a relatively stable performance region. Appendix~G provides paired Wilcoxon signed-rank tests across Accuracy, F1, Precision, and Recall, showing that UASPL achieves statistically significant improvements over many baseline methods.

\subsubsection{Ablation Studies}

We conduct three ablation settings to analyze the effectiveness of the main designs in UASPL. The first setting removes the evidential modeling of class probabilities, so that UASPL reduces to the corresponding traditional SPL baselines with point-estimate predictions. The other two settings retain the evidential modeling of class probabilities but replace the proposed uncertainty-dependent KL coefficient with two alternative KL-weighting schemes: 
(i) a fixed coefficient ($\beta=1$), which removes the uncertainty-dependent weighting and uses a constant KL weight throughout training; and
(ii) an annealed coefficient ($\beta_t=(t+1)/T$), which removes the uncertainty-dependent weighting and follows a standard linear schedule commonly used in evidential deep learning, where $T$ denotes the total number of self-paced learning rounds and $t$ is the current round index starting from 0. 
All three settings are evaluated under the hard, linear, and mixture regularizers.

\begin{table}[!ht]
\centering
\caption{Ablation results of UASPL on datasets in terms of average accuracy.}
\label{tab:ablation_accuracy_summary}
\fontsize{10pt}{12pt}\selectfont
\renewcommand{\arraystretch}{1.12}
\setlength{\tabcolsep}{4pt}
\begin{tabular}{>{\centering\arraybackslash}p{0.17\linewidth}>{\centering\arraybackslash}p{0.33\linewidth}ccc}
\toprule
Regularizer & Variant & Avg. Acc. & $\Delta$ vs. Full & Rank \\
\midrule
Hard & SPL baseline & 0.8143 & -0.0142 & 4 \\
Hard & UASPL ($\beta=1$) & 0.8228 & -0.0057 & 3 \\
Hard & UASPL ($\beta_t$) & 0.8253 & -0.0032 & 2 \\
Hard & Full UASPL & \textbf{0.8285} & \textbf{0} & \textbf{1} \\
\addlinespace[0.25em]
Linear & SPL baseline & 0.8255 & -0.0029 & 4 \\
Linear & UASPL ($\beta=1$) & 0.8267 & -0.0017 & 3 \\
Linear & UASPL ($\beta_t$) & 0.8272 & -0.0012 & 2 \\
Linear & Full UASPL & \textbf{0.8284} & \textbf{0} & \textbf{1} \\
\addlinespace[0.25em]
Mixture & SPL baseline & 0.8179 & -0.0107 & 4 \\
Mixture & UASPL ($\beta=1$) & 0.8246 & -0.0040 & 3 \\
Mixture & UASPL ($\beta_t$) & 0.8256 & -0.0030 & 2 \\
Mixture & Full UASPL & \textbf{0.8285} & \textbf{0} & \textbf{1} \\
\bottomrule
\end{tabular}
\end{table}

The ablation results are summarized in Table~\ref{tab:ablation_accuracy_summary}, and the corresponding detailed accuracy results on datasets are provided in Table D.17 of the Supplementary Material. Across the hard, linear, and mixture self-paced regularizers, the full UASPL achieves the highest average accuracy. Compared with the corresponding traditional SPL baselines, the variants with evidential modeling of class probabilities achieve higher average accuracy, suggesting that modeling class probabilities with a Dirichlet distribution is beneficial for self-paced learning. Furthermore, the full UASPL outperforms the fixed- and annealed-KL variants under all three regularizers, indicating that the proposed uncertainty-dependent KL coefficient is more effective than predefined KL-weighting schemes. These results support the effectiveness of both the evidential modeling of class probabilities and the uncertainty-dependent KL regularization in UASPL.

\subsubsection{Evaluation on Image Benchmarks}
To further assess the applicability of UASPL beyond  tabular datasets, additional experiments are conducted on four widely used image classification benchmarks, namely CIFAR-10, FashionMNIST, MNIST, and SVHN.

For the image datasets, deep backbone networks such as ResNet18 are adopted, and all reported methods are evaluated under matched experimental settings. For CLU and WSPLBF, we report only their EKNN-based variants, i.e., CLU1 and WSPLBF1, because preliminary results show that their ECM-based variants, i.e., CLU2 and WSPLBF2, exhibit similar performance trends but require substantially higher computational cost under deep image backbones. The backbone network is pretrained for 5 epochs using cross-entropy loss, and 5 Monte Carlo runs are conducted for each dataset. In addition, following \cite{2026generalized}, a correct-evidence regularization term is incorporated into Eq.~\eqref{tag:L_total_def} in the image experiments to address the optimization issues caused by low-evidence samples under deep architectures. Moreover, for image classification with deep backbones, we adopt a stage-dependent hybrid score for sample selection to better accommodate the training dynamics of deep networks:
\begin{equation}
\text{score}_i=(1-r_t)\,\mathrm{Norm}(\mathcal{L}_{\text{total}}^{(i)})+r_t\,\mathrm{Norm}(1-u_i),
\end{equation}
where $r_t=(t+1)/T$ denotes the stage progress ratio, with $T$ being the total number of self-paced learning rounds and $t$ the current round index starting from 0. Here, $\mathcal{L}_{\text{total}}^{(i)}$ reflects the current overall optimization status of the $i$-th sample, while $u_i$ characterizes whether the sample still lacks sufficient evidence and therefore retains further room for evidence updating. Accordingly, the early stages mainly favor samples with smaller total losses, i.e., samples that can already be handled relatively well by the current model. As $r_t$ increases, the selection criterion gradually places more emphasis on the evidence-related term, so that the model does not remain focused only on samples that have already become highly certain, but instead pays more attention to samples with relatively small losses yet still insufficient evidence. In this way, the selection preference gradually shifts from samples that are currently easy to optimize to samples that are learnable and still retain room for further evidence updating.

\begin{table*}[htbp]
\centering
\caption{Accuracy comparison across the four image datasets.}
\label{tab:image_accuracy_by_dataset}
\resizebox{\textwidth}{!}{
\begin{tabular}{>{\centering\arraybackslash}p{0.16\linewidth}cccc}
\toprule
\textbf{Method} & \textbf{CIFAR10} & \textbf{FashionMNIST} & \textbf{MNIST} & \textbf{SVHN} \\
\midrule
Direct & \underline{0.9398} $\pm$ \underline{0.0011} & 0.9470 $\pm$ \underline{0.0009} & 0.9957 $\pm$ \underline{0.0003} & 0.9602 $\pm$ 0.0014 \\
SPL & 0.9040 $\pm$ 0.0031 & 0.9340 $\pm$ 0.0014 & 0.9944 $\pm$ 0.0009 & 0.9539 $\pm$ 0.0022 \\
SPL\_linear & 0.9028 $\pm$ 0.0040 & 0.9324 $\pm$ 0.0044 & 0.9945 $\pm$ 0.0006 & 0.9562 $\pm$ 0.0024 \\
SPL\_mixture & 0.9001 $\pm$ 0.0034 & 0.9316 $\pm$ 0.0026 & 0.9910 $\pm$ 0.0018 & 0.9513 $\pm$ 0.0032 \\
CLU1 & 0.8165 $\pm$ 0.0111 & 0.8297 $\pm$ 0.1242 & 0.9866 $\pm$ 0.0036 & 0.8953 $\pm$ 0.0188 \\
WSPLBF1 & 0.9197 $\pm$ 0.0021 & 0.9371 $\pm$ 0.0014 & 0.9953 $\pm$ 0.0004 & 0.9610 $\pm$ 0.0011 \\
MW-Net & \textbf{0.9421} $\pm$ 0.0025 & \underline{0.9487} $\pm$ \textbf{0.0006} & \underline{0.9961} $\pm$ \textbf{0.0002} & \underline{0.9626} $\pm$ 0.0015 \\
active\_bias & 0.9397 $\pm$ 0.0023 & 0.9465 $\pm$ 0.0013 & 0.9957 $\pm$ 0.0004 & 0.9622 $\pm$ 0.0015 \\
SPL\_conf & 0.9140 $\pm$ 0.0019 & 0.9401 $\pm$ 0.0020 & 0.9949 $\pm$ 0.0004 & 0.9544 $\pm$ 0.0018 \\
SPL\_marg & 0.9162 $\pm$ 0.0028 & 0.9407 $\pm$ 0.0011 & 0.9949 $\pm$ 0.0005 & 0.9554 $\pm$ 0.0025 \\
MC\_MI & 0.9059 $\pm$ 0.0034 & 0.9316 $\pm$ 0.0013 & 0.9947 $\pm$ 0.0005 & 0.9556 $\pm$ \underline{0.0010} \\
MC\_pBALD & 0.9141 $\pm$ 0.0033 & 0.9363 $\pm$ 0.0020 & 0.9946 $\pm$ 0.0008 & 0.9612 $\pm$ 0.0013 \\
MC\_BalEnt & 0.9011 $\pm$ 0.0026 & 0.9350 $\pm$ 0.0015 & 0.9959 $\pm$ \underline{0.0003} & 0.9598 $\pm$ 0.0022 \\
SPL\_SPUP & 0.9039 $\pm$ 0.0024 & 0.9359 $\pm$ 0.0013 & 0.9944 $\pm$ 0.0009 & 0.9551 $\pm$ 0.0031 \\
UASPL & 0.9346 $\pm$ \textbf{0.0010} & \textbf{0.9491} $\pm$ 0.0010 & \textbf{0.9963} $\pm$ 0.0004 & \textbf{0.9653} $\pm$ \textbf{0.0006} \\
\bottomrule
\end{tabular}
}
\end{table*}

Due to space limitations, the main text reports only the Accuracy results on the four image datasets in Table~\ref{tab:image_accuracy_by_dataset}. The complete results for Accuracy, F1, Precision, and Recall on all compared methods are reported in Tables~H.24--H.27 of the Supplementary Material.

As shown in Table~\ref{tab:image_accuracy_by_dataset}, UASPL remains highly competitive across the four image benchmarks. On CIFAR-10, UASPL does not achieve the highest mean Accuracy, but it yields the smallest standard deviation, indicating stronger stability across repeated runs. On FashionMNIST, UASPL achieves the best mean Accuracy while maintaining a low standard deviation. On MNIST, where most methods already approach saturated performance, UASPL still attains the highest mean Accuracy. On SVHN, UASPL achieves both the best mean Accuracy and the smallest standard deviation. It is worth noting that MW-Net is a strong meta-learning-based reweighting baseline, but it maintains an additional weighting network and updates it using a small unbiased meta-data set, which usually makes it more time-consuming than non-meta-learning baselines under deep image backbones. The full four-metric results in Appendix~H also show that UASPL maintains competitive mean performance while achieving low variance across the image benchmarks.

\section{Conclusion}
\label{sec_conclusion}

This paper focuses on the issue that traditional self-paced learning methods mostly rely on the loss values to measure sample difficulty, whereas samples with small losses are not necessarily reliably simple samples for the model. To this end, we propose an Uncertainty-Aware Self-Paced Learning (UASPL) framework based on evidential neural networks, which integrates model-generated evidential uncertainty together with the label-fitting loss into the self-paced learning objective. Furthermore, UASPL incorporates an interpretable sample selection preference and demonstrates strong generality when embedded into different SPL variants. Extensive experiments verify that UASPL achieves superior classification performance, interpretability, and generalization compared with multiple baseline methods. 

Moreover, UASPL provides promising insights for studies on self-paced learning by showing how model-generated predictive reliability can be incorporated into sample selection strategies beyond loss-only criteria and externally introduced auxiliary information. As UASPL mainly focuses on reliability-aware sample-difficulty estimation in SPL, future work may proceed in the following directions: (1) investigating a tailored self-paced regularizer to more faithfully reflect sample importance by considering predictive reliability; (2) establishing a more comprehensive sample selection strategy that jointly takes sample diversity and uncertainty estimation into account; (3) developing an adaptive dynamic scheduling mechanism to further enhance the flexibility of UASPL in the sample selection process.

\section*{Acknowledgment}

The work is partially supported by National Natural Science Foundation of China (Grant No. 62303382, 62403388), and by Qin Chuangyuan high-level innovation and entrepreneurship talent program of Shaanxi (Grant No.QCYRCXM-2023-108), and by Guangdong Basic and Applied Basic Research Foundation (Grant No. 2023A1515110784), Shaanxi Fundamental Science Research Project for Mathematics and Physics (Grant No. 23JSQ034).

\bibliography{mybibfile}

\end{document}